\documentclass[letterpaper]{article} 
\usepackage{aaai23}  
\usepackage{times}  
\usepackage{helvet}  
\usepackage{courier}  
\usepackage[hyphens]{url}  
\usepackage{graphicx} 
\usepackage{natbib}  
\usepackage{caption} 
\usepackage[labelsep=period]{caption}

\usepackage{algorithm}
\usepackage{algpseudocode}
\usepackage{newfloat}
\usepackage{listings}
\DeclareCaptionStyle{ruled}{labelfont=normalfont,labelsep=colon,strut=off} 
\floatstyle{ruled}
\newfloat{listing}{tb}{lst}{}
\floatname{listing}{Listing}
\usepackage{tikz}
\usepackage{comment}
\usepackage{amsmath,amssymb} 
\usepackage{color}
\usepackage{tabto}

\usepackage{booktabs}
\usepackage{array}
\usepackage{longtable}
\usepackage{subcaption}
\usepackage[accsupp]{axessibility}  
\usepackage{amsmath}
\usepackage{amssymb}
\usepackage{upgreek}
\usepackage{tablefootnote}
\usepackage{mathtools}

\DeclareMathOperator*{\argmin}{arg\,min}

\title{POEM: Polarization of Embeddings for Domain-Invariant Representations}
\author {
	Sang-Yeong Jo,
	Sung Whan Yoon\thanks{Sung Whan Yoon is the corresponding author.}
}
\affiliations {
	Graduate School of Artificial Intelligence, Ulsan National Institute of Science and Technology (UNIST), Republic of Korea\\
	jsy7058@unist.ac.kr, shyoon8@unist.ac.kr
}

\usepackage{bibentry}

\begin{document}

\maketitle

\begin{abstract}
	Handling out-of-distribution samples is a long-lasting challenge for deep visual models. In particular, domain generalization (DG) is one of the most relevant tasks that aims to train a model with a generalization capability on novel domains. Most existing DG approaches share the same philosophy to minimize the discrepancy between domains by finding the domain-invariant representations. On the contrary, our proposed method called POEM acquires a strong DG capability by learning domain-invariant and domain-specific representations and polarizing them. Specifically, POEM co-trains category-classifying and domain-classifying embeddings while regularizing them to be orthogonal via minimizing the cosine-similarity between their features, i.e., the polarization of embeddings. The clear separation of embeddings suppresses domain-specific features in the domain-invariant embeddings. The concept of POEM shows a unique direction to enhance the domain robustness of representations that brings considerable and consistent performance gains when combined with existing DG methods. Extensive simulation results in popular DG benchmarks with the PACS, VLCS, OfficeHome, TerraIncognita, and DomainNet datasets show that POEM indeed facilitates the category-classifying embedding to be more domain-invariant.
\end{abstract}

\section{Introduction}

Despite the immense effort dedicated during the past decade, enhancing deep models to acquire a strong generalization capability on novel data distribution remains a daunting challenge.
For computer vision, particularly, the distributional shift of the image domain between the train and test sets, known as domain shift, provokes significant performance degradation of deep visual models. Domain generalization (DG), the task of interest here, pursues developing algorithmic methods to overcome the domain shift. 
Specifically, the DG task assumes that an image classification model is trained on the data from source domains, such as photos, sketches, cartoons, etc., then the model is tested on the target domains which are not shown in the training phase.

To overcome the domain shift problem, most of the existing DG approaches are built upon the philosophy of minimizing the discrepancy across source domains, which aims to obtain domain-invariant knowledge.
First of all, various algorithmic approaches have been proposed to minimize the divergence measurements across domains, such as 
the contrastive loss for alignment of in-class features from domains \cite{contrastive1,MASF}, the Kullback-Leibler divergence \cite{kullback1951information,MASF}, and the maximum mean discrepancy between domains \cite{MMD,MMD_DG}.
Another branch of approaches tries to utilize domain-specific information to learn domain-invariant representation via the employment of per-domain embedding network \cite{DSN} and domain classifiers \cite{AD_DA1}. 
Also, multi-task self-supervised learning \cite{multi-generalization,extrinsic_supervision_DG_MTL,jigsaw_DG_MTL}, optimization-based meta-learning \cite{MLDG, MASF}, and ensemble learning \cite{EoA,bestsource,ensemble1} are shown to enhance the model robustness across domain shifts.
On the other hand, another group of algorithms pursues to erase domain-related spurious factors in input space, such as the texture of images \cite{regular_1} or sensitive features in representation space \cite{RSC} to obtain domain-invariant features.

In the surge of various DG approaches to suppress discrepancy between domains, a work of \cite{Domainbed} reveals that, when a model is carefully trained, Empirical Risk Minimization (ERM) of \cite{ERM}, which is probably the simplest approach for training across multiple domains, outperforms the existing complicated DG methods.
After the surprising findings, many researchers have turned attention to developing particular optimizers that make models robust, rather than employing explicit ways to find domain-invariant representation.
For instance, recent approaches beat many prior works by combining ERM with model averaging methods for seeking flatter minima in loss landscape \cite{swa,SWAD}.
In addition, a very recent work of \cite{MIRO} maximizes the mutual information between a DG model and a pretrained oracle representation, rather than adopting a particular way to make the DG model more domain-invariant.

To the best of our knowledge, most of the existing DG methods aim to discard domain-specific information to reduce the divergence of representations between different domains or indirectly utilize domain-specific information to facilitate the acquisition of domain-invariant representations.
Moreover, recently suggested methods of \cite{SWAD,MIRO} overlook the effort for finding domain-invariant representation and focus on the robust-guaranteeing optimization methods of models.
We want to emphasize a significant difference between the strategy of prior work and how humans identify image categories across different domains.
For a given image, human recognizes the image category and domain together, and construct domain-invariant features based on the understanding of domain-specific features, i.e., human clearly acknowledges how a cartoon-based cat looks different from a photograph-based cat.
In contrast, none of the existing DG methods can explicitly identify both the domain-specific and domain-invariant features, and distinctively learn them to build domain-robust knowledge.

With this motivation, we propose a DG method called POEM that aims to learn both domain-invariant and domain-specific features which are clearly separated from each other.
Specifically, POEM employs two distinctive embeddings for the category and domain classification tasks, respectively, and zero-forces their cosine similarity to strengthen the clear discrimination between two embeddings.
POEM eventually forces two representations of category and domain classification tasks to be orthogonal, where one contains domain-invariant features for category classification and another one bears domain-specific features for domain classification; here, we call the process as \textit{polarization}.

We empirically show that POEM promotes the category-classifying embedding to be more domain-invariant. 
Also, we informally describe how POEM improves the generalization capability.
The concept of POEM with the disentangled domain-specific and domain-invariant representations enlightens a unique direction to further improve the performance of the existing DG methods. 
Extensive simulations on the popular DG benchmarks including PACS \cite{PACS}, VLCS \cite{VLCS}, OfficeHome \cite{officehome}, TerraIncognita \cite{terraincognita}, and DomainNet \cite{DomainNet} demonstrate that POEM yields a considerable gain when combined with the cutting-edge DG algorithms.

\if false
When POEM is combined with previous state-of-the-art DG approach, MIRO\cite{MIRO} and a recent DG method called Stochastic Weight Averaging Densely (SWAD) of \cite{SWAD} which finds flat minima, our experimental result shows considerable performance gain \textcolor{red}{and achieves state-of-the-art performance} on the popular DG benchmarks including PACS \cite{PACS}, VLCS \cite{VLCS}, OfficeHome \cite{officehome}, TerraIncognita \cite{terraincognita}, and DomainNet \cite{DomainNet}.
\fi

\if false
To break through the domain shift problem, promising methodologies have been proposed by utilizing domain alignment \cite{Invariant_Feature_Representation,contrastive1,KL1,MMD,Multi-Adversarial,representation_via_representations}, meta-learning \cite{MASF}, ensemble learning \cite{EoA,SWAD}, self-supervised learning \cite{multi-generalization,extrinsic_supervision_DG_MTL,jigsaw_DG_MTL}, and regularization strategies \cite{regular_1,RSC}. The common principle of the methodologies for DG is that they have attempted to improve the robustness by ignoring or erasing domain information.
Most existing DG approaches operate on the same principle to minimize the discrepancy between different domains when the model is trained for classifying image categories.

Most approaches utilize domain-specific information to reduce the divergence between the representations of different domains or train a network to classify category and domain together. 
However, there is a significant difference between the existing methods and how humans generalize image categories across domains. Human selective perception recognizes the image domain and image category together and it is aware of differences between image category and image domain. Unlike existing strategies with this motivation, \textcolor{blue}{
	our strategy surpresses an category-classifying embedding to extract domain-invariant representation by teaching both category-classifying network and domain-classifying network that even though equivalent input samples are fed into them, the outputs are separate.
}

\textcolor{blue}{To this end,} we propose POEM that enlightens a distinctive way to obtain a strong generalization capability by jointly training domain-invariant and domain-specific embeddings. In the joint training, POEM forces the embeddings to show the clear disentanglement, i.e., the zero-forced similarity between domain-invariant and domain-specific features. To be specific, POEM trains the \textit{super embedding} which is a union of the two distinctive embeddings where one is for classifying image categories and the other one is for classifying image domain. 
The normal component embeddings, called elementary embeddings, are based on separate models \textcolor{blue}{for separate own output labels} and \textit{super embedding} polarizes superset of feature spaces of elementary embeddings. The essential part of POEM is to adopt particular loss terms where one is the similarity loss between the features from the separate elementary embeddings and the other is the discrimination loss of the embeddings. 
The loss terms encourage each embedding space to be orthogonal and discriminative to each other so that the capabilities for classifying categories and domains are disentangled. After the joint training, we take the category-classifying elementary embedding to classify image categories as a distinctive domain-invariant representation that detaches the domain-variant features from itself. Note that POEM do not remove domain features but polarizes category-classifying features from domain-variant features.

When the domain-invariant representation empowered by POEM is evaluated on the target domains, we confirm that a considerable performance gain on the target domains is obtained. To the best of our knowledge, POEM is a unique approach that finds the domain-invariant representation by disentangling the category-related and domain-related information. We emphasize that the proposed method can easily be in conjunction with other existing DG methods.
When POEM is combined with previous state-of-the-art DG approach, MIRO\cite{MIRO} and a recent DG method called Stochastic Weight Averaging Densely (SWAD) of \cite{SWAD} which finds flat minima, our experimental result shows considerable performance gain on the popular DG benchmarks including PACS \cite{PACS}, VLCS \cite{VLCS}, OfficeHome \cite{officehome}, TerraIncognita \cite{terraincognita}, and DomainNet \cite{DomainNet}.

\fi

The main contributions of this paper are as follows:
\begin{itemize}
	\item We propose a method called POEM that enhances the DG capability via polarization of domain-invariant and domain-specific features.
	\item We provide a brief explanation that informally describes the improvement of DG ability based on the separation of domain-invariant and domain-specific features.
	\item We demonstrate a consistent and considerable performance gain of POEM when combined with the cutting-edge DG methods. 
\end{itemize}

\section{Related Work}
Beyond the brief summary of prior domain generalization (DG) methods in the Introduction, we herein focus on describing the highly-related works to POEM and the recent trend of DG algorithms.

\subsection{Aligning Domains via Domain-Specific Knowledge}
Most of the existing DG methods rely on the principle that minimization of the discrepancy across training domains improves the DG capability of models.
A group of methods in \cite{DSN, bestsource} adopts per-domain embeddings that classify categories of images in each domain, and reduce the discrepancy between them. 
As another strategy that utilizes domain-specific knowledge to acquire domain-invariant representation, the method in \cite{AD_DA1} employs a classifier of image domains and gradient-reversely co-trains it with the image category classifier. The process makes the model inept to recognize domains.
In contrast to the prior methods, our method POEM explicitly co-trains category- and domain-classifying embeddings and disentangles them to achieve better generalization, which is never been proposed.
The methods of \cite{DSN, bestsource} does not employ domain-classifying representations, and the algorithm of \cite{AD_DA1} adopts just a domain classifier, not a domain-classifying embedding.


\if false
\subsection{Adopting Multi-task Learning}
Another branch of approaches considers the multi-domain setting of the DG task as the multi-task learning scenario. 
Prior methods of \cite{extrinsic_supervision_DG_MTL, jigsaw_DG_MTL} train a multi-task learning task with the union of the DG task and self-supervised learning task to enhance the DG performance. 
The work of \cite{multi-generalization} trains a model over a set of multiple self-supervised learning tasks to prepare a better initialization of the DG task.
Some notable algorithms in \cite{MLDG, MASF} borrow the concept of meta-learning to prepare a well-generalized model for unseen domains.
\fi


\subsection{Erasing Domain-dependancy}
On the other hand, DG approaches with the domain-erasing strategy pursue to discard domain-dependent features. 
A method called NGLCM of \cite{regular_1} regularizes domain-dependent texture features of images extracted by Gray-Level Co-occurrence Matrix (GLCM) of \cite{GLCM1,GLCM2}. 
Representation Self-Challenging (RSC) of \cite{RSC} learns to mask sensitive features in the representation space, which are believed to be domain-dependent.
Common Specific Decomposition (CSD) of \cite{decomposition} decomposes the model parameters into the common and domain-specific parts to identify the domain-invariant model parameters.
When compared to the domain-erasing methods, POEM erases domain-dependent parts from domain-invariant representations by reducing the similarity between the category- and domain-classifying embeddings. However, POEM is fundamentally different from the method of \cite{regular_1} that relies on visual characteristics such as texture, and the methods of \cite{RSC, decomposition} that are not able to recognize explicit domain-dependent representations.

\subsection{Optimizing Models for Generalization}
After the authors of \cite{Domainbed} claim that Empirical Risk Minimization (ERM) of \cite{ERM} shows outperforming performance beyond the existing complicated DG methods, ensemble learning of moving average models (EoA) of \cite{EoA} shows the improved DG performance by just averaging model parameters during the ERM training steps.
A group of approaches surpasses combines ERM and the model averaging methods that find flatter minima in loss space \cite{swa,SWAD}.
POEM is also built upon ERM, which is the simplest way to handle the DG task and is easily plugged in with the flat-minima searching methods called Stochastic Weight Averaging Densely (SWAD) of \cite{SWAD} for cutting-edge DG performance.
As the MIRO case, POEM can enhance the domain invariance of a model in conjunction with SWAD, which pays less attention to finding domain-invariant features.

\subsection{Utilizing Pretrained Knowledge}
Well-pretrained models from other datasets can be used for better DG performance.
As a very recent work, Mutual Information Regularization with Oracle (MIRO) of \cite{MIRO} aims to maximize the mutual information between the pretrained oracle representation and the target model's representations for better generalization.
MIRO does not adopt an explicit way to find domain-invariant features but just makes a model be similar to the oracle.
Our method is essentially different from MIRO, so POEM can be in conjunction with MIRO to yield an additional performance gain via enhancing the domain invariance.


\section{Proposed Method}
In this section, the problem settings of domain generalization (DG) are presented and the details of the proposed algorithm POEM are described.

\subsection{Problem Settings of Domain Generalization}
Let us denote the set of training domains as $\mathcal{D}=\{\mathcal{D}_{k}\}_{k=1}^{K}$ where $D_{k}$ is the $k$-th training domain. 
For a classification model $f(x;\theta)$ and the loss function $\mathcal{L}$, the objective of the DG task is to find the model parameter $\theta$ which is generalized well on the target domain $\mathcal{T}$, i.e.,
\begin{equation}
	\theta^{*} = \argmin_{\theta}{\mathcal{L}\big(f(\mathbf{x};\theta),y \: ; \mathcal{D} \big)},
\end{equation}
where $(\mathbf{x},y)$ is a pair of input and class label from $\mathcal{T}$.

\subsection{Model Description of POEM}
POEM consists of a set of \textit{elementary embeddings}. For the DG task, POEM contains two elementary embeddings, one is for image category classification, and the other one is for image domain classification. Here, we extend the concept to contain $N$ number of elementary embeddings for a more general description. 
Based on the architecture, POEM adopts \textit{disentangling loss} for spatially separating the elementary embeddings and \text{discrimination loss} for discriminating the features from different embeddings.


\textbf{Set of elementary embeddings:}
Let us denote the set of elementary embedding as $\mathfrak{F}: \mathbb{R}^{D}\rightarrow\mathbb{R}^{N\times L}$ which is the set of elementary embeddings $\mathfrak{F}=\{f_{i}\}_{i=1}^{N}$ with model parameter $\mathrm{\Theta}=\{\theta_{i}\}_{i=1}^{N}$:
\begin{equation}
	\mathfrak{F}(\mathbf{x}\:;\mathrm{\Theta}) \triangleq \big\{ f_{i}(\mathbf{x}\:;\theta_{i})\big\}_{i=1}^{N},
\end{equation}
where $N$ is the number of elementary embeddings.
Each elementary embedding $f_{i}$ that is parameterized by $\theta_{i}$ maps an input $\mathbf{x}$ to the feature vector with the length of $L$.
For the set of elementary embeddings, there exist $N$ elementary tasks with different classifiers, i.e., category classifiers and domain classifiers for the DG task.
The classifier $\boldsymbol{\mathrm{\Phi}}$ is the set of $N$ classifiers for elementary tasks.
For a given input $\mathbf{x}$ and $i$-th elementary embedding, the classification loss $\mathcal{L}_{c}$ is calculated with cross-entropy $\mathcal{H}$ with the probability from the Softmax computation and target label $y^{(i)}$:
\begin{equation}\label{eq:elementary_loss}
	\mathcal{L}_{c}^{(i)}(\mathbf{x},y) = \mathcal{H}\Big(\text{Softmax}\big\{ f_{i}(\mathbf{x}\:;\theta_{i})\mathrm{\Phi}_{i} \big\}, y^{(i)}\Big)
\end{equation}
For the DG task, there exist $N=2$ pairs of elementary embedding and classifier for category and domain classification, respectively.
For instance, the PACS dataset contains seven categories, three train domains, and a single target domain. 
POEM then contains two elementary embeddings that classify seven categories and three domains for each.

\textbf{Disentangling loss:}
POEM computes disentangling loss for separating elementary embeddings from each other.
To be specific, the cosine-similarity loss between features from different embeddings is zero-forced.
For a given input $\mathbf{x}$, the disentangling loss $\mathcal{L}_{s}^{(i,j)}(x)$ for a pair of $i$ and $j$-th  elementary embeddings is calculated as follows:
\begin{equation} \label{eq:disent_loss}
	\mathcal{L}_{s}^{(i,j)}(\mathbf{x}) = \lvert K\big( f_{i}(\mathbf{x}\:;\theta_{i}), f_{j}(\mathbf{x}\:;\theta_{j}) \big)\rvert,
\end{equation}
where $K(\cdot,\cdot)$ is the cosine similarity function of two vectors. The absolute operation $\lvert \cdot \rvert$ is for making the similarity be positive. We select cosine similarity for the disentangler to orthogonalize two embedded features. 

\textbf{Discrimination loss:}
POEM adopts discrimination loss which is to recognize the index of embeddings for a given feature. The discriminator $\mathbf{W}$ is a simple classifier with $N$ classification weights: $\mathbf{W} = \{{w_{i}}\}_{i=1}^{N}$. 
For a given $\mathbf{x}$ and $i$-th elementary embedding, discrimination loss $\mathcal{L}_{d}^{(i)}(\mathbf{x})$ is computed with cross-entropy with the probability from Softmax calculation and target label $i$:
\begin{equation} \label{eq:discrim_loss}
	\mathcal{L}_{d}^{(i)}(\mathbf{x}) = \mathcal{H}\Big( \text{Softmax}\big\{ f_{i}(\mathbf{x};\theta_i)\mathbf{W} \big\}, i\Big)
\end{equation}
For the DG case, the discrimination is a binary classification to figure out the index of the embedding from the input feature vector. 

\begin{figure}[t]
	\centering
	\includegraphics[width=0.49\textwidth,trim={1cm 2cm 0cm 1cm}]{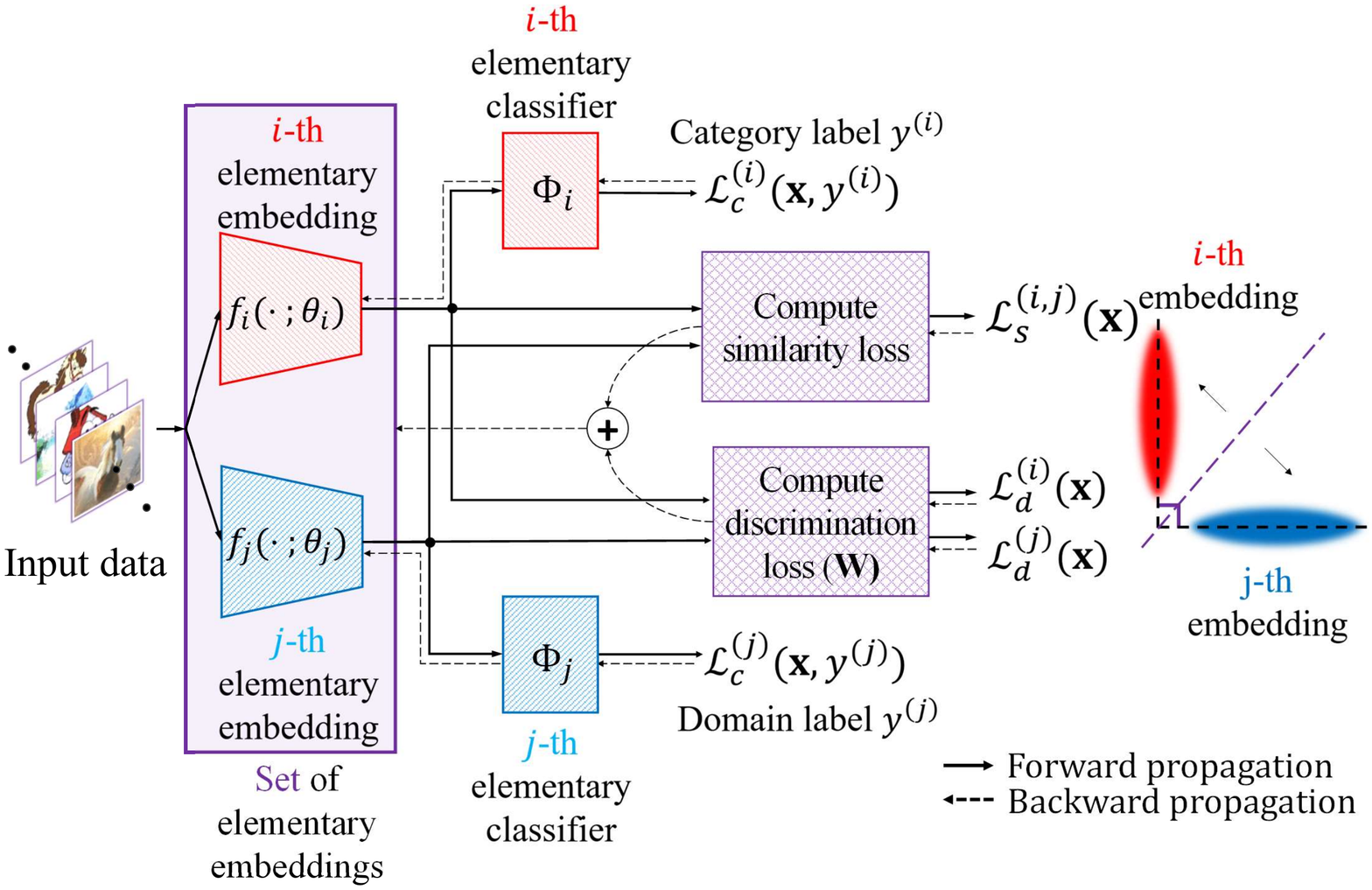}
	\caption{Proposed model architecture of POEM}
	\label{fig:overview}
\end{figure}	

In Fig. \ref{fig:overview}, the model architecture of POEM for the DG task is illustrated. The set of elementary embeddings contains two elementary embeddings $f_{i}$ and $f_{j}$ for the category-classification and the domain-classification tasks, respectively. Based on the two classfiers $\mathrm{\Phi}_i$ and $\mathrm{\Phi}_j$ for classifying image categories and domains as respectively, POEM calculates two classification loss terms denoted as $\mathcal{L}_{c}$. For orthogonalizing features from two elementary embeddings, POEM computes the disentangling loss $\mathcal{L}_{s}$. For the final loss term, a discriminator with parameter $\mathbf{W}$ calculates the discrimination loss $\mathcal{L}_{d}$. 

\subsection{Learning Procedures of POEM}
\textbf{Training phase:}
The learning procedures of POEM are based on the most straightforward framework called Empirical Risk Minimization (ERM) \cite{ERM,Domainbed} that minimizes the empirical risk, which is the average of category-classification losses $\mathcal{L}$ over the source domains.
The empirical risk is formulated as follows:
\begin{equation}
	\mathcal{\hat{E}}_{\mathcal{B}}(\theta) \triangleq \frac{1}{|\mathcal{B}|}\sum_{(\mathbf{x},y)\in\mathcal{B}}\mathcal{L}(f(\mathbf{x};\theta),y),
\end{equation}
where $\mathcal{B}=\{\mathcal{B}_{k}\}_{k=1}^{K}$ is a mini-batch, and $\mathcal{B}_{k}$ is a sampled mini-batch from  $\mathcal{D}_{k}$ of domain $k$. $f(\cdot\:;\theta)$ is an embedding parameterized by $\theta$, and $y$ is the image category label.
Similarly, POEM trains learnable parameters including $\mathrm{\Theta}$, $\mathrm{\Phi}$ and $\mathbf{W}$ to minimize the empirical risk as follows:
\begin{equation}
	\mathcal{\hat{E}}_{\mathcal{B}}(\mathrm{\Theta}, \mathrm{\Phi}, \mathbf{W}) \triangleq \frac{1}{|\mathcal{B}|}\sum_{(\mathbf{x},y)\in\mathcal{B}}\mathcal{L}(\mathfrak{F}(\mathbf{x};\mathrm{\Theta}), \mathrm{\Phi}, \mathbf{W}, y).
\end{equation}
The particular loss term $\mathcal{L}$ is computed by considering the classification loss of elementary tasks $\mathcal{L}_{c}$, the disentangling loss $\mathcal{L}_{s}$ between different embeddings, and the discrimination loss $\mathcal{L}_{d}$ for each embedding which are aforementioned:
\begin{equation}
	\begin{split}
		&\mathcal{L}(f\big(\mathbf{x};\mathrm{\Theta}), \mathrm{\Phi}, \mathbf{W}, y\big) = \\
		&\frac{1}{N}\displaystyle\sum_{i=1}^{N}\Big\{ \mathcal{L}_{c}^{(i)}(\mathbf{x},y) + \mathcal{L}_{d}^{(i)}(\mathbf{x}) + \sum_{j\neq i}^{N}\mathcal{L}_{s}^{(i,j)}(\mathbf{x}) \Big\}.
	\end{split}
\end{equation}
Then the set of parameters $\mathrm{\Theta}$, $\mathrm{\Phi}$ and $\mathbf{W}$ are updated by computing the gradients of the empirical risk, i.e., $\mathcal{\hat{E}}_{\mathcal{B}}(\mathrm{\Theta}, \mathrm{\Phi}, \mathbf{W})$:
\begin{equation} \label{eq:gradient}
	\begin{split}
		&\nabla\mathcal{\hat{E}}_{\mathcal{B}}(\mathrm{\Theta}, \mathrm{\Phi}, \mathbf{W}) = \frac{1}{N|\mathcal{B}|} \sum_{(\mathbf{x},y)\in\mathcal{B}} \sum_{i=1}^{N}\nabla  \mathcal{L}(\mathfrak{F}\big(\mathbf{x};\mathrm{\Theta}), \mathrm{\Phi}, \mathbf{W}, y\big) \\
	\end{split}
\end{equation}

\textbf{Testing phase:}
In testing, POEM keeps the embedding and classifier for the category-classifying task but drops other embeddings and classifiers. 
With the retained embedding and classifier, i.e., $f_{z}(\cdot\:;\theta_{z})$ and $\mathrm{\Phi}_{z}$, POEM is evaluated on the samples in the target domains $\mathcal{T}$, where $z$ is the index of the elementary embedding for classifying categories of images.
Algorithm \ref{alg:POEM} presents the pseudocode of POEM.

\newcommand{\factorial}{\ensuremath{\mbox{\sc Factorial}}}
\begin{algorithm}
	\caption{Training procedures for POEM}\label{euclid}
	\textbf{Input:} Training domain $\mathcal{D}$, Number of elementary embeddings $N$, learning rate $\eta$ \\
	\textbf{Initialization:} Initial weights $\mathrm{\Theta}_{0}$, $\boldsymbol{\mathrm{\Phi}}_{0}$, and $\mathbf{W}_{0}$, set of elementary embeddings $\mathfrak{F}(\cdot\:;\mathrm{\Theta}_{0}$)  \\
	\textbf{Output:} Parameterized model $f_{z}(\cdot\:;\theta_z)$ and classifier $\mathrm{\Phi}_{z}$
	\begin{algorithmic}[1]
		\For{$\tau=1,\cdots,T$}
		\State Sample a mini-batch $\mathcal{B}=\{\mathcal{B}_{k}\}_{k=1}^{N}$, where $\mathcal{B}_{k}\in\mathcal{D}_{k}$
		\For{$(\mathbf{x},y)\in\mathcal{B}$} 
		\State Set $\mathcal{L}_{c}= 0$, $\mathcal{L}_{s}= 0$, and $\mathcal{L}_{d}= 0$
		\For{$i=1,\cdots,N$}
		\State $\mathcal{L}_{c} \leftarrow \mathcal{L}_{c} + \mathcal{L}_{c}^{(i)}(\mathbf{x},y)$ \Comment{Eq. \eqref{eq:elementary_loss}}
		\State $\mathcal{L}_{d} \leftarrow \mathcal{L}_{d} + \mathcal{L}_{d}^{(i)}(\mathbf{x})$ \Comment{Eq. \eqref{eq:discrim_loss}}
		\For{$j=1,\cdots,N$}
		\If{$j\neq i$}
		\State $\mathcal{L}_{s} \leftarrow \mathcal{L}_{s} + \mathcal{L}_{s}^{(i,j)}(\mathbf{x})$ \Comment{Eq. \eqref{eq:disent_loss}}
		\EndIf
		\EndFor
		\EndFor
		\EndFor
		\State $\mathcal{\hat{E}}_{\mathcal{B}} \leftarrow \frac{1}{N|\mathcal{B}|}(\mathcal{L}_{c} + \mathcal{L}_{s} +\mathcal{L}_{d})$ 
		\State ($\mathrm{\Theta}, \boldsymbol{\mathrm{\Phi}}, \mathbf{W}) \leftarrow (\mathrm{\Theta}, \boldsymbol{\mathrm{\Phi}}, \mathbf{W}) - \eta\nabla{\mathcal{\hat{E}}_{\mathcal{B}}}$
		\EndFor
		\State \textbf{Return} $f_{z}(\cdot\:;\theta_{z})$ and $\mathrm{\Phi}_{z}$, where $z$ is the index of the category-classifying embedding
	\end{algorithmic}
	\label{alg:POEM}
\end{algorithm}





\begin{table*}[!ht]
	\centering
	\begin{tabular}{cccccc|c}
		\toprule
		\textbf{Method} & PACS & VLCS & OfficeHome & TerraInc & DomainNet & Average\\
		\midrule
		MMD \cite{MMD_DG} & 84.7 & 77.5 & 66.4 & 42.2 & 23.4 & 58.8\\
		Mixstyle \cite{Mixstyle} & 85.2 & 77.9 & 60.4 & 44.0 & 34.0 & 60.3\\
		GroupDRO \cite{GroupDRO} & 84.4 & 76.7 & 66.0 & 43.2 & 33.3 & 60.7\\
		IRM \cite{IRM} & 83.5 & 78.6 & 64.3 & 47.6 & 33.9 & 61.6\\
		ARM \cite{ARM} & 85.1 & 77.6 & 64.8 & 45.5 & 35.5 & 61.7\\
		VREx \cite{VREx} & 84.9 & 78.3 & 66.4 & 46.4 & 33.6 & 61.9\\
		CDANN \cite{CDANN} & 82.6 & 77.5 & 65.7 & 45.8 & 38.3 & 62.0\\
		DANN \cite{DANN} & 83.7 & 78.6 & 65.9 & 46.7 & 38.3 & 62.6\\
		RSC \cite{RSC} & 85.2 & 77.1 & 65.5 & 46.6 & 38.9 & 62.7\\
		MTL \cite{MTL} & 84.6 & 77.2 & 66.4 & 45.6 & 40.6 & 62.9\\
		I-Mixup \cite{imixup} & 84.6 & 77.4 & 68.1 & 47.9 & 39.2 & 63.4\\
		MLDG \cite{MLDG} & 84.9 & 77.2 & 66.8 & 47.8 & 41.2 & 63.6\\
		SagNet \cite{segnet} & 86.3 & 77.8 & 68.1 & 48.6 & 40.3 & 64.2\\
		CORAL \cite{CORAL} & 86.2 & 78.8 & 68.7 & 47.7 & 41.5 & 64.5\\
		SWAD \cite{SWAD} & 88.1 & 79.1 & 70.6 & 50.0 & 46.5 & 66.9\\
		MIRO \cite{MIRO} & 85.4 & 79.0 & 70.5 & 50.4 & 44.3 & 65.9\\
          ERM$^\dagger$ \cite{ERM} & 84.1 $\pm$ 0.7& 77.9 $\pm$ 0.8 & 67.0 $\pm$ 0.3 & 46.8 $\pm$ 1.1 & \textbf{44.1} $\pm$ 0.0 & 64.0\\
		\textbf{POEM} (Ours) & \textbf{86.7} $\pm$ 0.2 & \textbf{79.2} $\pm$ 0.6 & \textbf{68.0} $\pm$ 0.2 & \textbf{49.5} $\pm$ 0.6 & 44.0 $\pm$ 0.0 & \textbf{65.5} ($\uparrow$ 1.5\%)\\
		\bottomrule
		SWAD$^\dagger$ \cite{SWAD} & 88.3 $\pm$ 0.3 & 77.7 $\pm$ 0.3 & \textbf{70.7} $\pm$ 0.1 & 49.7 $\pm$ 0.6 & 46.2 $\pm$ 0.0 & 66.5\\
		\textbf{SWAD$^\dagger$ + POEM} (Ours) & \textbf{88.5} $\pm$ 0.2 & \textbf{79.4} $\pm$ 0.3 & 70.5 $\pm$ 0.1 & \textbf{51.5} $\pm$ 0.1 & \textbf{47.1} $\pm$ 0.0 & \textbf{67.4} ($\uparrow$ 0.9\%)\\
		\hline
		MIRO$^\dagger$ \cite{MIRO} & 85.4 $\pm$ 0.3 & 79.1 $\pm$ 0.7 & 70.7 $\pm$ 0.0 & \textbf{49.7} $\pm$ 0.2 & 44.3 $\pm$ 0.2 & 65.8\\
		\textbf{MIRO$^\dagger$ + POEM} (Ours) & \textbf{86.7} $\pm$ 0.4 & \textbf{79.1} $\pm$ 0.2 & \textbf{71.4} $\pm$ 0.0 & \textbf{49.3} $\pm$ 0.8 & \textbf{44.3} $\pm$ 0.2 & \textbf{66.1} ($\uparrow$ 0.3\%)\\
		\bottomrule
		MIRO + SWAD$^\dagger$ & 87.7 $\pm$ 0.3 & 78.5 $\pm$ 0.3 & 71.3 $\pm$ 0.1 & 51.0 $\pm$ 0.2 & 46.9 $\pm$ 0.0 & 67.1\\
		\textbf{MIRO + SWAD$^\dagger$ + POEM} (Ours) & \textbf{88.5} $\pm$ 0.1 & \textbf{79.5} $\pm$ 0.3 & \textbf{71.7} $\pm$ 0.1 & \textbf{51.6} $\pm$ 0.0 & \textbf{47.1} $\pm$ 0.0 & \textbf{67.7} ($\uparrow$ 0.6\%)\\
		\hline
	\end{tabular}
 	\footnotesize{$^\dagger$ indicates our reproduced experiments based on the DomainBed settings. $\uparrow$ indicates the performance gains obtained by POEM.} 
    \caption{Domain generalization accuracies on the five benchmarks}
	\label{tab:target_performance}
\end{table*}

\subsection{Understanding of POEM}
Herein, we informally explain how POEM improves the domain generalization capability. 
Although the explanation is not a formal mathematical analysis, we conceptually understand how the elementary embeddings of POEM are constructed and how the well-trained POEM achieves an improved generalization capability beyond ERM. 

Before the explanation, let us introduce some useful notations.
We denote the trained set of elementary embeddings of POEM as $\mathfrak{F}(\cdot;\Theta^{*})=\{f_{i}(\cdot;\theta_{i}^{*})\}_{i=1}^{N}$ where $f_i(\cdot;\theta_i^{*})$ is $i$-th elementary embedding with the learned parameters $\theta_{i}^{*}$, and $N$ is the number of elementary embeddings. $N_{i}$ is the number of labels for the classification task of $i$-th embeddings, e.g., when we have seven image categories and four domains, $N_{1}=7$ and $N_{2}=4$.
$\mathcal{X}$ is the input distribution that contains input samples $\mathbf{x}$.
Let us denote the distribution of feature vectors of $i$-th elementary embedding as $\mathcal{Z}_{i}^{*}$.
Based on the notations, let us describe the following desirable properties of the trained POEM embeddings.

\textbf{Property 1.} \textit{(from the discrimination loss $\mathcal{L}_{d}^{(i)}$)} When the feature $\mathbf{z}_{i}^{*}$ is extracted by  $i^{th}$ embedding, i.e., $\mathbf{z}_{i}^{*} \sim \mathcal{Z}_{i}^{*}$, then 
\begin{equation}
\mathbf{z}_{i}^{*}\cdot\mathbf{w}_{i} \ge \displaystyle\max_{j\neq i}(\mathbf{z}_{i}^{*}\cdot\mathbf{w}_{j}).
\end{equation}

Based on the discrimination loss, POEM is trained to identify the index of embedding where a given feature is extracted. Thus the property is desirable. 
POEM tries to separate the feature distribution of each embedding so that the distributions are not overlapped.

\textbf{Property 2.} \textit{(from the disentangling loss $\mathcal{L}_{s}^{(i,j)}$)} When two feature vectors are extracted from different  $i^{th}$ and $j^{th}$ embeddings for a single input $\mathbf{x}$, then
\begin{equation}
\big|K\big(f_{i}(\mathbf{x};\theta_{i}^*), f_{j}(\mathbf{x};\theta_{j}^*)\big)\big| \simeq 0.
\end{equation}

Based on the disentangling loss for a given input, POEM is trained to minimize the cosine similarity between two features that are extracted from different embeddings. Thus the property is also desirable. 

With Property 1, the distributions of embeddings are separated but not orthogonalized. On the other hand, with Property 2, the sample-wise orthogonalization is guaranteed but the distributions can be overlapped.
When POEM tries to achieve these both properties, the feature distributions of different embeddings should be separated and orthogonalized, i.e., the polarization of embeddings.
In the following section, we visually show the separation of feature distributions of different embeddings, and empirically confirm the zero-forced cosine similarity values between randomly-sampled pair of features from different embeddings. 

Based on the understanding of POEM, we informally provide the following claim to explain how POEM achieves the improved generalization capability. 
First, let us process the singular value decomposition (SVD) of the matrix $\mathbf{M}_{j}$ formed by the collected feature vectors from $j^{th}$ embedding, i.e., $\mathbf{M}_{j}=\mathbf{U}_{j}\boldsymbol{\Sigma}_{j}\mathbf{V}_{j}^{T}$. Then let us project a feature vector $\mathbf{z}_{i}^{*}$ from different $i^{th}$ embedding to the vector space $\mathbf{U}_{j}\boldsymbol{\Sigma}_{j}$.
Then the power of the projected feature vector will be zero-forced because the dominant components of $\mathbf{U}_{j}$ would be orthogonal to $\mathbf{z}_{i}^{*}$ due to the polarization of embeddings.

\textbf{Claim 1.} \textit{(Information separation of embeddings)} When feature vector $\mathbf{z}_{i}^{*}\sim \mathcal{Z}_{i}^{*}$ is projected to the space formed by the features from different $j^{th}$ embedding, then the power of the projected feature is minimized to zero:
\begin{equation}
||\mathbf{z}_{i}^{*}\mathbf{U}_{j}\boldsymbol{\Sigma}_{j}||^{2} \simeq 0.
\end{equation}

It implies the information separation between embeddings, i.e., for the DG task, the features for the domain-classifying embedding are zero-forced in the category-classifying embedding space.
In other words,  features from the category-classifying embedding are domain-invariant, or do not contain the information for domain-classification. Otherwise, the domain-specific features contained in the category-classifying features will remain non-zero when projected to the domain-classifying embedding. 
The formal analysis of POEM remains as a future work.

\if false
\subsection{Theoretical Analysis of POEM}	
Herein, we theoretically guarantee the improved domain generalization capability.
Before our theoretical analysis, let us introduce some useful notations.
We denote the trained set of elementary embeddings of POEM as $\mathfrak{F}(\cdot;\Theta^{*})=\{f_{i}(\cdot;\theta_{i}^{*})\}_{i=1}^{N}$ where $f_i(\cdot;\theta_i^{*})$ is $i$-th elementary embedding with the learned parameters $\theta_{i}^{*}$, and $N$ is the number of elementary embeddings. $N_{i}$ is the number of labels for the classification task of $i$-th embeddings, e.g., when we have seven image categories and four domains, $N_{1}=7$ and $N_{2}=4$.
$\mathcal{X}$ is the input distribution that contains input samples $\mathbf{x}$.
Let us denote the distribution of feature vectors of $i$-th elementary embedding as $\mathcal{Z}_{i}^{*}$.

Based on the notations, let us introduce the following assumptions which are originated from the loss terms in POEM, i.e., per-elementary embedding classification task, discrimination loss, and disentangling loss.

\textbf{Assumption 1.} \textit{(Bounded $L_{2}$-norm)} For all $i\in[N]$, $L_{2}$-norm of embedded vectors are upper bounded by $B^{2}$:
\begin{equation}
	||f_{i}(\mathbf{x};\theta_{i}^{*})||_{2} \leq B^{2},
\end{equation}
, where $\mathbf{x}\sim\mathcal{X}$.

\textbf{Assumption 2.} \textit{(Bounded classifiers)} For all $i\in[N]$ and $k\in[N_{i}]$, $L_2$-norm of classifier $\mathbf{w}_{i}$ for the discrimination loss and task classifier $\boldsymbol{\phi}_{i}^{k}$ are upper bounded by $B^{2}$:
\begin{equation}
	||\mathbf{w}_{i}||_{2} \leq B^{2} \:\: \text{ and } \:\: ||\boldsymbol{\phi}_{i}^{k}||_{2} \leq B^{2}.
\end{equation}

\textbf{Assumption 3.} \textit{(from the discrimination loss $\mathcal{L}_{d}^{(i)}$)} For all $i\in[N]$, the inner product between  $\mathbf{z}_{i}^{*} \sim \mathcal{Z}_{i}^{*}$ and the corresponding discriminator weight $\mathbf{w}_{i}$ satisfies the following inequality with probability $1-\epsilon$, where $\delta$, $\epsilon<1$ are positive real numbers:
\begin{equation}
	\mathbb{P}_{\mathcal{Z}_{i}^*}\Big[\mathbf{z}_{i}^{*}\cdot\mathbf{w}_{i}-\displaystyle\max_{j\neq i}(\mathbf{z}_{i}^{*}\cdot\mathbf{w}_{j}) \ge B^{2}(1-\delta) \Big]=1-\epsilon.
\end{equation}

\textbf{Assumption 4.} \textit{(from the classification loss $\mathcal{L}_{c}^{(i)}$)} For all $i\in[N]$, the feature vector $f_i(\mathbf{x};\theta_{i}^{*})$ satisfies the following inequality with probability $1-\epsilon$, where $\delta$, $\epsilon<1$ are positive real numbers:
\begin{align}
	\mathbb{P}_{\mathcal{X}}\Big[f_{i}(\mathbf{x};\theta_{i}^{*})\cdot\boldsymbol{\phi}_{i}^{k} - & \log\sum_{l\neq k} \exp{\big(f_{i}(\mathbf{x};\theta_{i}^{*})\cdot\boldsymbol{\phi}_{i}^{l}\big)} \\
	& \ge B^{2}(1-\delta) \Big] = 1-\epsilon,
\end{align}
where $k$ is the label of $\mathbf{x}\sim \mathcal{D}$ is for $i$-th task, and $\delta$, $\epsilon<1$ are positive real numbers.

\textbf{Assumption 5.} \textit{(from the disentangling loss $\mathcal{L}_{s}^{(i,j)}$)} For any different indices $i, j\in[N]$, the absolute of cosine similarity between two features from $i$-th and $j$-th elementary embeddings is upper bounded by $\gamma$ with probability $1-\eta$, where $\gamma$, $\eta<1$ are arbitrary small positive real numbers.
\begin{equation}
	\mathbb{P}_{\mathcal{X}}\Big[ \big|K\big(f_{i}(\mathbf{x};\theta_{i}^*), f_{j}(\mathbf{x};\theta_{j}^*)\big)\big| \le \gamma \Big] = 1-\eta,
\end{equation}
, where $\mathbf{x}\sim \mathcal{D}$.

With these assumptions, Lemma 1 guarantees the separations of different elementary embeddings.

\textbf{Lemma 1.} \textit{(Separation of elementary embeddings)} For the features of $i$-th and $j$-th elementary embeddings, the expected inner product is upper bounded by $\delta B^{2}$ with probability $1-\alpha$, where $\alpha$, $\delta < 1$ are positive real numbers.
\begin{equation}
	\mathbb{P}_{(\mathcal{Z}_{i}^*,\mathcal{Z}_{j}^*)}\Big[ |\mathbf{z}_{i}^{*}\cdot\mathbf{z}_{j}^*| \le \delta B^{2} \Big] = 1-\alpha, 
\end{equation}
where $\mathbf{z}_{i}^{*} \sim \mathcal{Z}_{i}^{*}$ and $\mathbf{z}_{j}^{*} \sim \mathcal{Z}_{j}^{*}$.

Before quantifying the information separation between embeddings, let us introduce a definition as follows:

\textbf{Definition 1.} \textit{(Principal components of elementary embeddings)} For $i$-th elementary embedding, singular value decomposition of matrix $\mathbf{M}_{i}$ which is the collection of sufficiently many sampled features, is $\mathbf{M}_{i}=\mathbf{U}_{i}\boldsymbol{\Sigma}_{i}\mathbf{V}_{i}^{T}$.

\textbf{Theorem 1.} \textit{(Information separation of elementary embeddings)} For $i$-th elementary embedding, the L2-norm of the feature $\mathbf{z}_{i}^{*}\sim\mathcal{Z}_{i}^{*}$ that is informative for task $i$ is upper bounded by $B\alpha^{2}$ with probability $1-\beta$, when projected on $\mathbf{M}_{j}$, where $j\neq i$ and $j\in[N]$:
\begin{equation}
	\mathbb{P}_{\mathcal{Z}_{i}^{*}}\Big[ \big|\big|\mathbf{z}_{i}^{*}\mathbf{M}_{j}\big|\big|_{2} \leq B\alpha^{2} \Big] = 1-\beta.
\end{equation}

Theorem 1 says that a feature vector from an elementary embedding is zero-forced when projected to the space spanned by different elementary embeddings. It implies the information separation between embeddings, i.e., for the DG task, the features for the domain-classifying embedding are zero-forced in the category-classifying embedding space.

Before mentioning the last mathematical argument, let us define $f(\cdot;\theta)$ as a single embedding for classifying image categories. 
Here, we claim that the POEM's elementary embedding $f(\cdot;\theta^{*}_{i})$ shows a tight generalization bound than the separately trained category-classifying embedding $f(\cdot;\theta)$, where $i$ is the index of category-classification embedding of POEM.
We introduce generalization bound from Input Compression Bound of \cite{IBN} and the domain divergence formulation of \cite{SWAD}.
Let us denote the true loss on target domain $\mathcal{T}$ and the empirical loss on source domain $\mathcal{D}$ of the POEM-based category-classifying embedding as $\mathcal{E}_{\mathcal{T}}^{*}$ and $\hat{\mathcal{E}}^{*}_{\mathcal{D}}$. 
Also, let us denote the empirical loss of the ERM-based category-classifying embedding $f(\cdot;\theta)$ as $\hat{\mathcal{E}}_{\mathcal{D}}$. $\mathcal{Z}$ is the feature distribution of $f(\cdot;\theta)$.

\textbf{Theorem 2.} \textit{(Tighter generalization bound)} For POEM-based category-classifying embedding $f(\cdot;\theta^{*}_{i})$, following generalization bound is hold with probability $1-\zeta$:
\begin{align}
	\mathcal{E}^{*}_{\mathcal{T}} &< \hat{\mathcal{E}}^{*}_{\mathcal{D}} + \frac{1}{2K}\sum_{k=1}^{K}\textbf{Div}(\mathcal{D}_{k},\mathcal{T}) + \sqrt{\frac{2^{I(X;\mathcal{Z}_{i}^{*})} + \log(2/\zeta)}{m}} \nonumber\\
	& \leq  \hat{\mathcal{E}}_{\mathcal{D}} + \frac{1}{2K}\sum_{k=1}^{K}\textbf{Div}(\mathcal{D}_{k},\mathcal{T}) + \sqrt{\frac{2^{I(X;\mathcal{Z})} + \log(2/\zeta)}{m}},
\end{align}
where $K$ is the number of source domains and $\textbf{Div}(\mathcal{P},\mathcal{Q}) \triangleq 2\sup_{A}|\mathbb{P}_{\mathcal{P}}(A) - \mathbb{P}_{Q}(A)|$ is a divergence between $\mathcal{P}$ and $\mathcal{Q}$, and $I(\mathcal{X},\mathcal{Y})$ is the mutual information between $\mathcal{X}$ and $\mathcal{Y}$. It implies that POEM achieves a tighter generalization bound than ERM. Proofs are given in Supplementary.

\fi

\section{Experimental Results}
\subsection{Experiment Settings}
\textbf{Benchmarks:} We have conducted extensive experiments to evaluate POEM on the five popular domain generalization (DG) benchmarks based on PACS \cite{PACS} (containing 9,991 images, 7 classes and 4 domains), VLCS \cite{VLCS} (containing 10,729 images, 5 classes, and 4 domains), OfficeHome \cite{officehome} (containing 15,588 images, 65 classes, and 4 domains), TerraIncognita \cite{terraincognita} (containing 24,788 images, 10 classes, and 4 domains), and DomainNet \cite{DomainNet} (containing 586,575 images, 345 classes, and 6 domains). 
For each benchmark, if a domain is selected as the target domain, then the remaining domains are designated to be the training source domains. 
We test all cases for each target domain and take the average of accuracies.
\if false
The hardware of our experiments is based on NVIDIA Quadro RTX 8000, 400GB RAM, and Xeon(R) Gold 5218R CPU @ 2.10GHz. The simulation environment for the experiments is on Ubuntu 20.04 with python 3.8.12, pytorch 1.7.1, torchvision 0.8.2, CUDA 11.0. 
\fi
Our experiments are run on the DomainBed framework of \cite{Domainbed}, which is publicly released under the MIT license to evaluate the existing DG methods\footnote{Code is available at github.com/JoSangYoung/Official-POEM}.
We follow the training and evaluation protocols of DomainBed of \cite{Domainbed}.
Also, we follow the data splitting introduced by the work of SWAD \cite{SWAD}. 

\textbf{Experiments Details:} We set the number of training iterations of POEM to be the same as the experiments done in \cite{SWAD}, i.e., PACS: 5,000, VLCS: 5,000, OfficeHome: 5,000, TerraIncognita: 5,000, DomainNet: 15,000 iterations.
When POEM is combined with MIRO of \cite{MIRO}, twice number of iterations are used, i.e., PACS: 10,000, VLCS: 10,000, OfficeHome: 10,000, TerraIncognita: 10,000, DomainNet: 30,000. 
For every elementary embedding, we adopt the ResNet50 architecture of \cite{resnet50} which is pretrained on the ImageNet dataset \cite{ImageNet} with freezing batch normalization parameters.
A mini-batch contains 32 images from each source domain in benchmark datasets. 
Due to the lack of memory in our simulation, a mini-batch for the DomainNet case contains 20 images for each source domain.
For all benchmarks, we have searched proper hyperparameters that include learning rates, dropout ratios, and weight decay rates for both elementary embeddings. 
Details of the hyperparameter values and the optimizers are described in Supplementary. 


\textbf{Methods to be considered:} Similar to other cutting-edge algorithms, POEM is built upon the ERM framework of \cite{ERM}. We denote the vanilla version of our method based on ERM as \textbf{POEM}. 
Also, the concept of POEM can be plugged in with other approaches.
We evaluate \textbf{SWAD + POEM}, \textbf{MIRO + POEM}, and \textbf{MIRO + SWAD + POEM}, by combining POEM with the most promising DG approaches.
POEM contains two elementary embeddings where one is for category-classifying, and the other is for domain-classifying.
SWAD + POEM adopts the optimization process for finding flat minima only for category-classifying embedding of POEM.
MIRO + POEM employs the pretrained oracle network to maximize the mutual information between the features from the oracle and both elementary embeddings of POEM. MIRO + SWAD + POEM combines all three methods. We described details of hyperparameters for SWAD and MIRO in Supplementary.

\if false
When it is trained with SWAD, the hyperparameters of optimum patient parameter $N_s$, an overfitting patient parameter $N_e$, and the tolerance rate $r$ for category-classifying embedding is set to 3, 6, 1.3, repectively. In the case of VLCS, the tolerance rate $r$ is set to 1.2, following the hyperparameter setting of SWAD\cite{SWAD}. For combination with MIRO, The hyperparameter of regularization coefficeint $\lambda$ for category-classifying embedding based on MIRO\cite{MIRO} is set to 0.1 for PACS and VLCS, 0.01 for others. The Hyperparameters for domain-classyfing embedding based on MIRO is searched in the range of [0, 0.1, 0.01] for each dataset. $\lambda$ of 0 indicates that domain-classifying embedding is set to ERM. For more details about hyperparameters, See Appendix.
For all cases, Adam optimizer is used in updating learnable parameters. In the test phase, category-classifying elementary embedding and classifier are picked, while the domain-task elementary embedding is discarded for computational complexity.
\fi


\subsection{Performance on Target Domain}
In Table \ref{tab:target_performance}, the DG performance of POEM, SWAD + POEM, MIRO + POEM, MIRO + SWAD + POEM are compared with the existing methods.
The accuracies are obtained by taking the averages over three trials.
We emphasize that POEM yields consistent performance gains when combined with ERM, SWAD, and MIRO.
Specifically, POEM obtains the averaged gains by +1.5\% beyond ERM and by +0.9\% beyond SWAD. 
Also, POEM yields an extra gain by +0.6\% beyond MIRO + SWAD.
The results confirm that POEM enlightens a unique way to enhance the domain-invariance of representations beyond cutting-edge algorithms. Performance on source domains are presented in Supplementary.

\if false
POEM based on the ERM framework shows the averaged performance gain by 1.5\% beyond ERM. When SWAD based on ERM is combined with POEM, the performance gain is 0.9\% beyond SWAD. Also, MIRO + POEM, and MIRO + SWAD + POEM shows averaged 
We emphasize that POEM always yields the performance gain over the accuracies of ERM, MIRO and SWAD with our implementations.
Notably, a considerable accuracy improvement is acquired for the VLCS dataset case, i.e., ERM $\rightarrow$ POEM: 77.88\% $\rightarrow$ 79.24\%, SWAD $\rightarrow$ SWAD + POEM: 77.72\% $\rightarrow$ 79.43\%.
\fi

\if false
\subsection{Performance on Source Domains}
We also measure the performance of POEM on the source domains. For each source domain, we split the dataset into 80\% of the samples for the training set and 20\% for the validation set. For evaluating the image category classification performance of POEM, the elementary embedding to classify image categories is used. For evaluating the domain classification performance, secondary embedding to classify image domains is used.
The averaged accuracies of POEM for classifying image categories and domains over three trials are shown in Table \ref{tab:source_performance}. We confirm the slight performance gain for all experiment cases. It implies POEM did not harm elementary embedding and the additional performance gain is effected by learning on super embedding rather than reducing training error of elementray embedding in source domain.
\begin{table*}[!htb]
	\centering
	\begin{tabular}{c}
		Accuracies for classifying image categories\\
		\begin{tabular}{cccccc|c}
			\toprule
			Algorithm & PACS & VLCS & OfficeHome & TerraInc & DomainNet & Average\\
			\midrule
			ERM$^\dagger$ & 96.99 $\pm$ 0.1 & 86.21 $\pm$ 0.1 & 80.38 $\pm$ 0.1 & 91.63 $\pm$ 0.1 & 60.00 $\pm$ 0.1 & 83.31\\
			\midrule
			\textbf{ERM + POEM$^\dagger$} & 97.14 $\pm$ 0.1 & 86.91 $\pm$ 0.1 & 80.85 $\pm$ 0.04 & 92.16 $\pm$ 0.1 & 60.82 $\pm$ 0.1 & 83.58\\
			\midrule
		\end{tabular}\\
		Accuracies for classifying image domains\\
		\begin{tabular}{cccccc|c}
			\toprule
			Algorithm & PACS & VLCS & OfficeHome & TerraInc & DomainNet & Average\\
			\midrule
			ERM$^\dagger$ & 99.02 $\pm$ 0.1 & 94.10 $\pm$ 0.1 & 85.13 $\pm$ 0.1 & 99.95 $\pm$ 0.02& 89.26 $\pm$ 0.1 & 93.49\\
			\midrule
			\textbf{ERM + POEM$^\dagger$} & 98.98 $\pm$ 0.1 & 93.99 $\pm$ 0.1 & 85.53 $\pm$ 0.1 & 99.96 $\pm$ 0.02& 89.32 $\pm$ 0.1 & 93.56\\
			\hline
		\end{tabular}
	\end{tabular}
    \caption{Validation accuracies on source domains}
	\footnotesize{\\$^\dagger$ indicates our implementation}\\
	\label{tab:source_performance}
\end{table*}

\subsection{Learning Trend of Similarity and Discrimination Loss Terms}
The learning trend of the key loss terms, which are the similarity loss $\mathcal{L}_{s}$ and the discrimination loss $\mathcal{L}_{d}$ are illustrated in Fig. \ref{fig:Training_Loss_of_POEM}. As presented in Fig. \ref{fig:loss_s}, averaged the cosine similarity between the category embedding and the domain embedding over VLCS domains decreases rapidly (See the solid line colored by red). It means that the feature vectors from the two elementary embeddings become orthogonal. In contrast, when we drop the similarity loss term in the training of POEM, the averaged cosine similarity between elementary embeddings over VLCS domains is not zero-forced (See the dotted line colored by blue). It infers the non-orthogonality between the features from elementary embeddings. 
Fig. \ref{fig:loss_d} shows the learning trend of the discrimination loss for the five benchmarks. The suppressed loss values mean that the features from different embeddings become distinctive. 

\begin{figure}
	\captionsetup{justification=centering}
	\centering
	\begin{subfigure}[hbt]{0.48\textwidth}
		\centering
		\includegraphics[width=\textwidth]{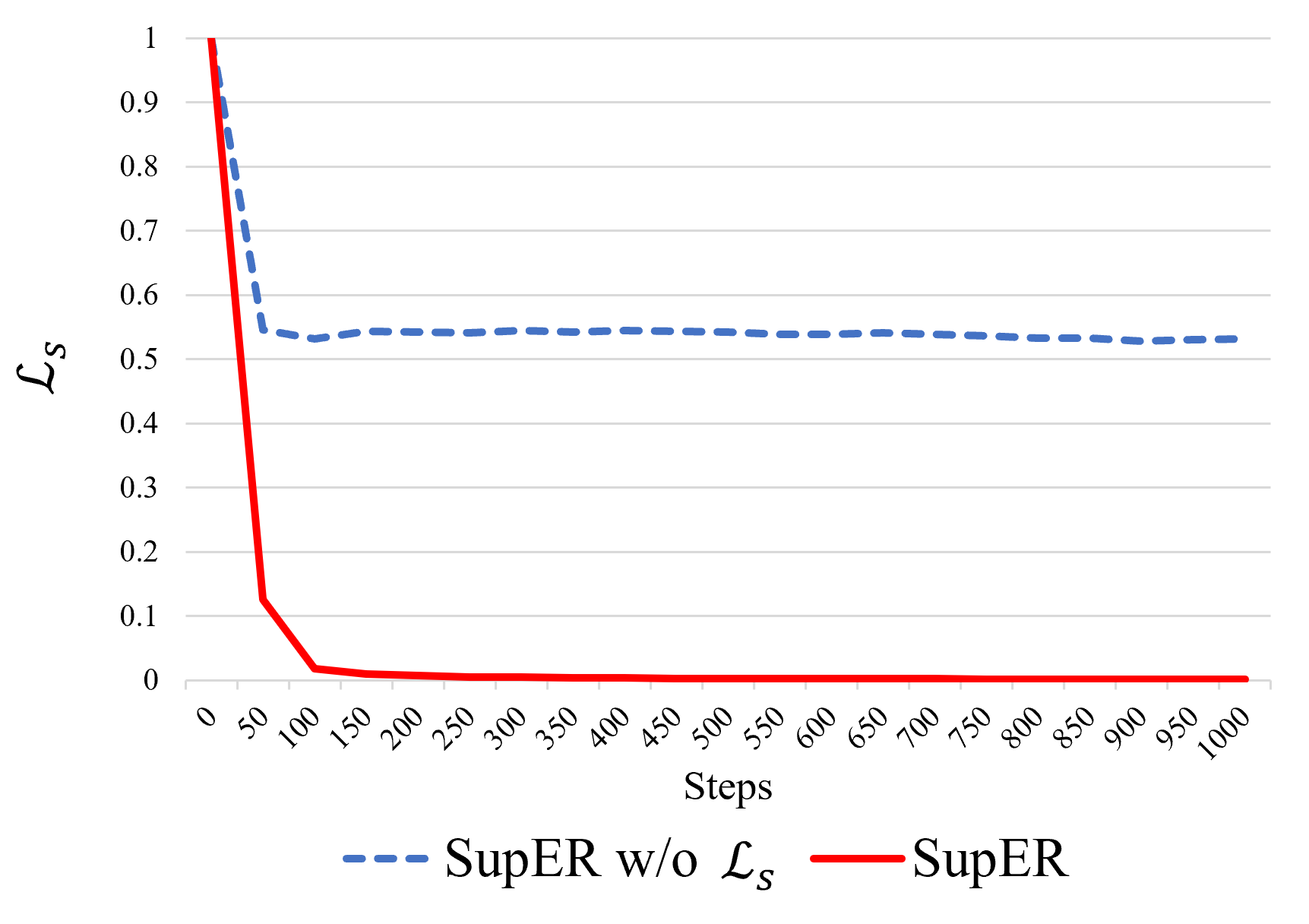}
		\caption{Trainig similarity loss $\mathcal{L}_s$}
		\label{fig:loss_s}
	\end{subfigure}
	\\
	\begin{subfigure}[hbt]{0.48\textwidth}
		\centering
		\includegraphics[width=\textwidth]{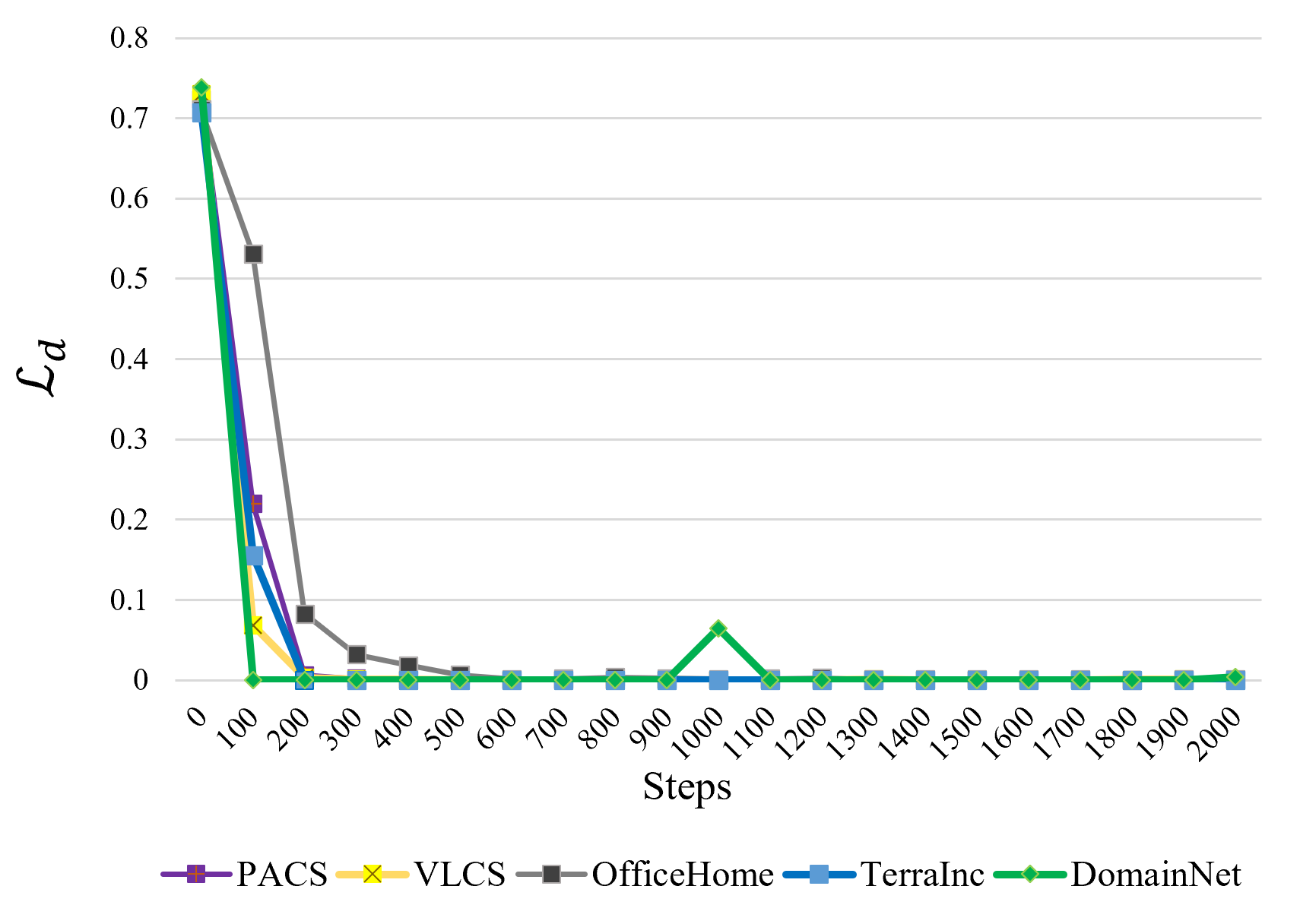}
		\caption{Training discriminator loss $L_d$}
		\label{fig:loss_d}
	\end{subfigure}
	\caption{Training losses of POEM}
	\label{fig:Training_Loss_of_POEM}
\end{figure}
\fi

\subsection{t-SNE Visualization of Embeddings}
To visualize the orthogonality between elementary embeddings, the t-SNE analysis of \cite{tsne} is conducted. 
We consider the experiment case of the VLCS benchmark where the target domain is the `SUN09' domain. 
Fig. \ref{fig:tSNE} is the t-SNE plot of features from the category-classifying embeddings and the domain-classifying embedding, which are colored by red and blue, respectively. 
This visualization clearly shows that POEM separates elementary embeddings without any overlaps.

\begin{figure}
	\captionsetup{justification=centering}
	\centering
	\includegraphics[width=0.4\textwidth,trim=0.3cm 0 0.3cm 0]
        {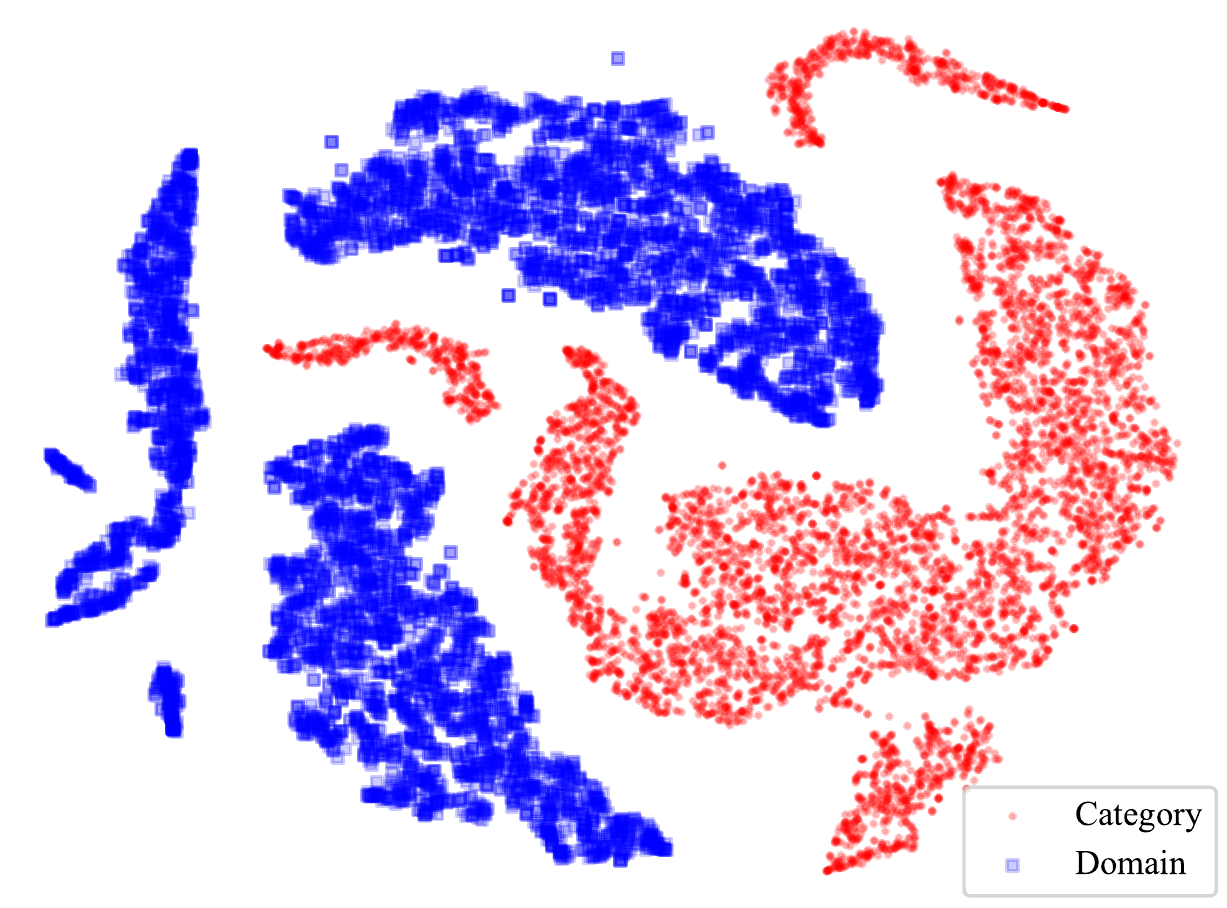}
	\caption{Visualization of features with embedding labels}
	\label{fig:tSNE}
\end{figure}

\if false
To visualize the disentanglement between the category and domain embeddings of POEM, the t-distributed stochastic neighbor embedding (t-SNE) analysis of \cite{tsne} is conducted. We consider the experiment case of the VLCS benchmark where the target domain is set to be the `SUN09' domain.

Fig. \ref{fig:tSNE_ERM} is the t-SNE plot of the category features with domain label from ERM. Fig. \ref{fig:tSNE_POEM} visualizes the category features with domoain label from ERM + POEM. When we focus on the LabelMe and VOC2007 domains with green and blue colors respectively, then the domains are more discriminative in the t-SNE plot of ERM. In contrast, the domains are slightly more mixed up in the t-SNE plot of ERM + POEM. It implies that the category features from POEM are more domain-invariant than the category features from ERM. 
\fi			

\if false
\begin{figure*}
	\captionsetup{justification=centering}
	\centering
	\begin{subfigure}[b]{0.4\textwidth}
		\centering
		\includegraphics[width=\textwidth]{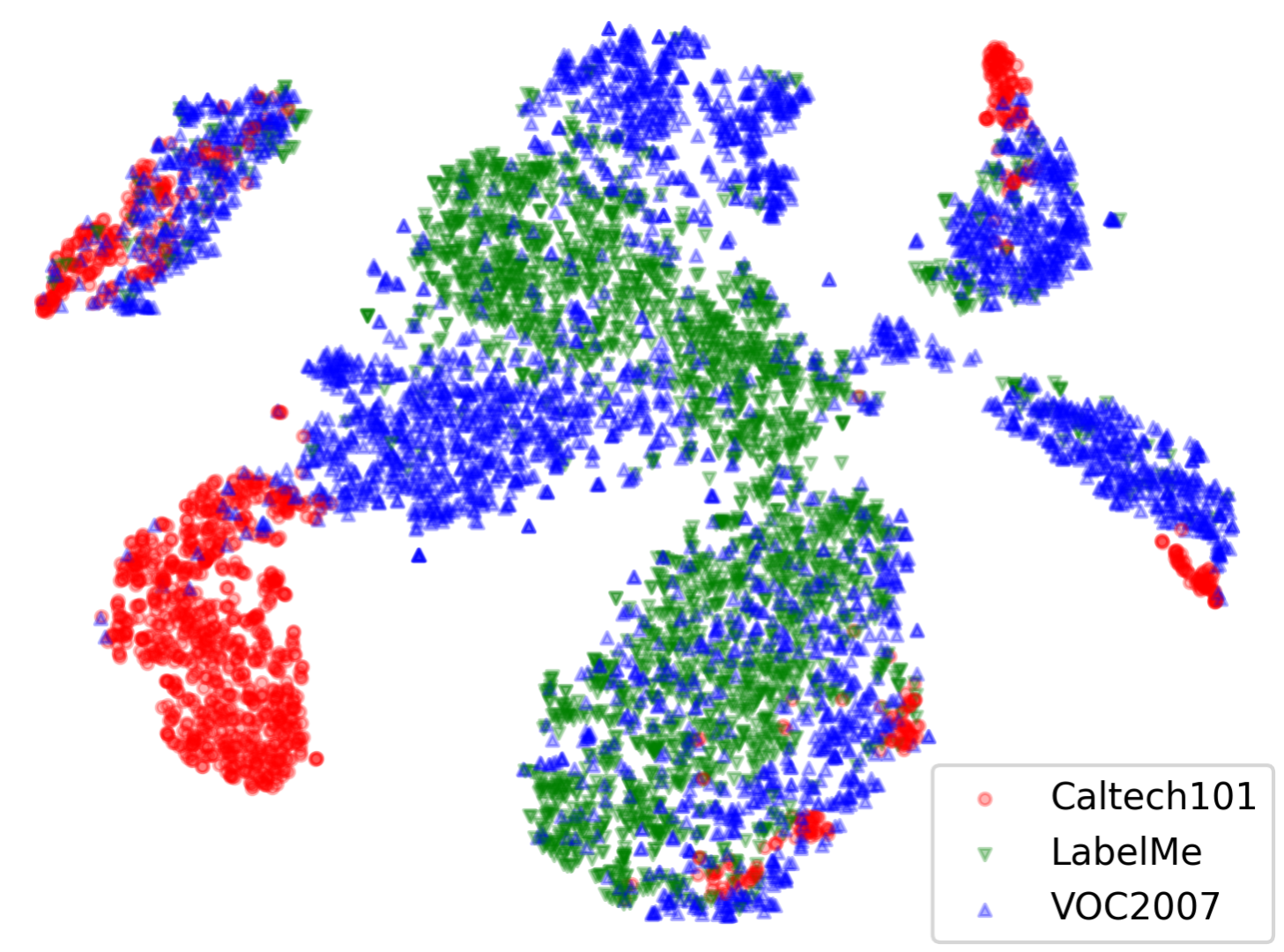}
		\caption{ERM-based features}
		\label{fig:tSNE_ERM}
	\end{subfigure}
	\begin{subfigure}[b]{0.4\textwidth}
		\centering
		\includegraphics[width=\textwidth]{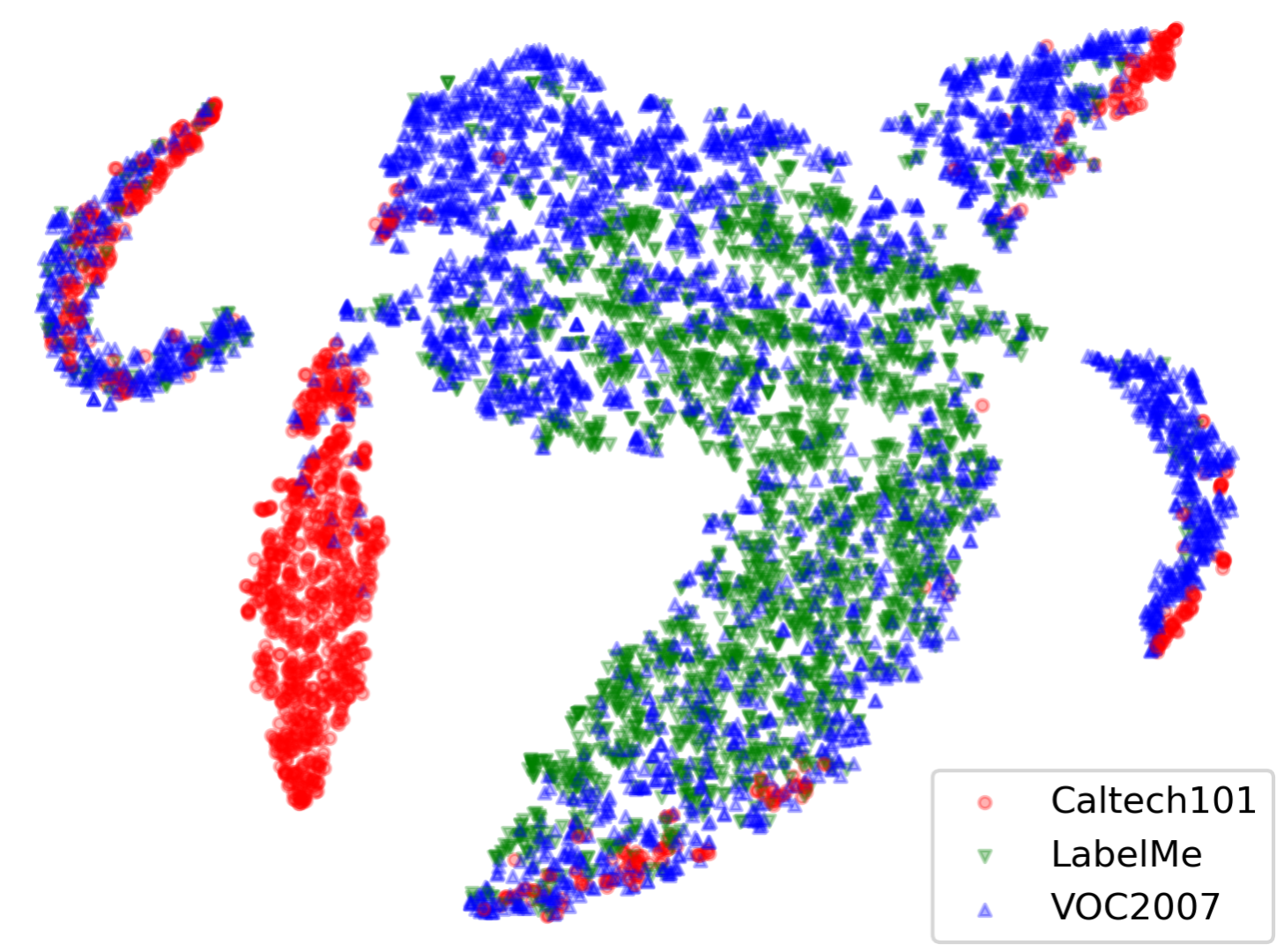}
		\caption{POEM-based features from the category embedding}
		\label{fig:tSNE_POEM}
	\end{subfigure}
	\caption{Visualization of the features from the category-classification embedding}
	\label{fig:tSNE}
\end{figure*}
\fi

\begin{table*}[!ht]
	\centering
	\begin{tabular}{cccccc|c}
		\toprule
		\textbf{Method} & PACS & VLCS & OfficeHome & TerraInc & DomainNet & Average\\
		\midrule
  		ERM$^\dagger$ & 84.09 $\pm$ 0.7 & 77.88 $\pm$ 0.8 & 67.00 $\pm$ 0.3 & 46.78 $\pm$ 1.1 & 44.13 $\pm$ 0.0 & 64.0\\
		POEM only with $\mathcal{L}_s$ & 84.86 $\pm$ 0.3 & 78.29 $\pm$ 0.5 & 67.12 $\pm$ 0.3 & 47.21 $\pm$ 2.0 & 43.78 $\pm$ 0.2 & 64.3\\
		POEM only with $\mathcal{L}_d$ & 84.96 $\pm$ 0.3 & 78.28 $\pm$ 0.4 & 67.45 $\pm$ 0.4 & 47.82 $\pm$ 0.8 & \textbf{44.04} $\pm$ 0.1 & 64.5\\
		\textbf{POEM} & \textbf{86.73} $\pm$ 0.3 & \textbf{79.24} $\pm$ 0.6 & \textbf{67.96} $\pm$ 0.2 & \textbf{49.48} $\pm$ 0.6 & \textbf{44.03} $\pm$ 0.0 & \textbf{65.5}\\
		\hline
	\end{tabular}
 \\ \footnotesize{$^\dagger$ indicates our implementation}
    \caption{Effect of loss functions in our method based on ERM over three trials}
	\label{tab:ablation}
\end{table*}

\subsection{Entropy Analysis of Embeddings}
For quantifying the domain-invariance of category-classifying features, we calculate the cross-entropy values when category-classifying features are used to classify domains. 
For the category embedding of POEM, the classifiers for domains are not prepared so we compute the domain-wise centroids $\{\mathbf{c}_{k}\}_{k=1}^{N}$ of features and utilize them as the classifiers for domains. 
After obtaining the domain centroids, the cross-entropy loss is calculated by measuring the probability based on the Euclidean distance between feature vectors and centroids, i.e.,
\begin{equation}
	P(y = k \: | \: \mathbf{x}) = \frac{\exp\big(-d(f_z(\mathbf{x}\:;\theta_z), \mathbf{c}_k)\big)}{\sum_{l=1}^{N} \exp\big(-d(f_{z}(\mathbf{x}\:;\theta_z), \mathbf{c}_{l})\big)},
	\label{equ:dist_prob}
\end{equation}
where $N$ is the number of source domains, $d(\cdot,\cdot)$ means the Euclidean distance, and $z$ is the index of the category embedding.
In addition, we train the ERM-based model on the same source domains and compute the cross-entropy loss with the same way. 
As shown in Table \ref{tab:entropy}, the category features from POEM show higher cross-entropy values when compared to the values of ERM. 
It indicates that POEM discards the domain-related information from the category embedding. OfficeHome, TerraIncognita, DomainNet is denoted as OH, Terra, DN, due to the space limit.

\begin{table*}[!ht]
	\centering
	\begin{tabular}{cccccc|c}
		\toprule
		\textbf{Method} & PACS & VLCS & OH & Terra & DN & Avg\\
		\midrule
		ERM & 0.22 & 0.27 & 0.09 & 0.14 & 0.06 & 0.16 \\
		\textbf{POEM} & 3.8e-05 & 1.0e-04 & 1.5e-04 & 2.9e-04 & 1.4e-03 & 3.94e-04 \\
		\hline
	\end{tabular}
    \caption{Averaged cosine similarity between category-classifying features and domain-classifying features}
	\label{tab:orthogonality}
\end{table*}

\begin{table}[!ht]
	\centering
	\begin{tabular}{cccccc|c}
		\toprule
		\textbf{Method} & PACS & VLCS & OH & Terra & DN & Avg\\
		\midrule
		ERM & 2.65 & 2.80 & 1.13 & 1.85 & 0.81 & 1.85 \\
		\midrule
		\textbf{POEM} & 2.98 & 4.01 & 1.41 & 2.12 & 0.68 & 2.24 \\
		\hline
	\end{tabular}
    \caption{Averaged cross-entropy for classifying domains with category-classifying features in 5 benchmark datasets}
	\label{tab:entropy}
\end{table}

\subsection{Orthogonality Analysis of Embeddings}
To confirm the orthogonality of different elementary embeddings of POEM, we compute the averaged cosine similarity values by randomly sampling two features from category- and domain-classifying embeddings.
Table \ref{tab:orthogonality} shows averaged cosine similarities in 5 benchmark datasets, by considering more than 1,000 samples for each domain.
As a counterpart, we prepare the ERM model for classifying image categories, and also prepare a separate ERM model that classifies image domains across different categories.
Then the averaged cosine similarity values are computed in the same way as POEM cases.
OfficeHome, TerraIncognita, DomainNet are denoted as OH, Terra, DN, respectively. 
The result shows that POEM makes elementary embeddings more orthogonal when compared to ERM for all benchmarks.
Note that ERM shows larger cosine similarities on the PACS and VLCS cases. 
By zero-forcing the cosine similarities, POEM indeed shows more considerable gains in that benchmarks when compared to others, as reported in Table \ref{tab:target_performance}.

\subsection{Ablation Analysis}
We conduct ablation studies of the loss terms of POEM. 
Table \ref{tab:ablation} shows the performance gain in the addition of the proposed loss functions. 
POEM only with $\mathcal{L}_{s}$ makes the cosine similarity between two paired features from a single image be zero. The performance gain for POEM only with $\mathcal{L}_{s}$ is +0.3\% when compared to ERM. The gain is quite small because the loss term $\mathcal{L}_{s}$ cannot separate the clusters of features from two embeddings.
Only with the discrimination loss $\mathcal{L}_d$, a moderated performance gain by +0.5\% is obtained beyond ERM.
However, the gain is not yet considerable because the loss cannot make two elementary embeddings orthogonal.
Finally, POEM with both loss terms eventually separates two embeddings in two orthogonal directions so that the considerable performance gain is achieved, i.e., +1.5\% beyond ERM.


\subsection{Complexity Analysis}
POEM prepares two elementary embeddings, but once training is over, POEM drops the domain-classifying embedding and utilizes only the category-classifying embedding for inference. It means that POEM shows the same level of memory and computational costs during testing when compared to ERM.
When we compare POEM with SWAD of \cite{SWAD}, which is a promising DG method, SWAD is required to store an additional moving average model during iterations.
It means that SWAD requires twice the number of parameters during the training phase, i.e., the same as the costs of POEM. 
MIRO of \cite{MIRO} shows the same level of costs as ERM during training, but MIRO requires additional costs for the pretraining of the oracles.

\section{Conclusion}
For achieving the robustness of the deep visual models on the out-of-distribution problem, we propose a method called POEM with a set of elementary embeddings where the elementary embeddings are trained to be disentangled with each other. 
We show that considerable performance gains can be achieved by combining POEM with other cutting-edge DG methods, including ERM, SWAD, and MIRO. 


\section{Discussion}
POEM is possibly extended to the more complicated generalization scenarios.
For example, the medical image classification task may include a variety of dimensions such as diseases, organs, patients, and types of imaging equipment. 
Then POEM with an embedding for each dimension possibly handles the generalization tasks across multiple dimensions. We leave it as a future work.
Specifically, we expect that POEM enables training the disease-related embedding invariant to the other factors, i.e., patients or medical imaging equipment. 
When considering the detection task for road objects, images would be diverse during daytime and night-time. 
By employing the day/night-classifying embedding, the concept of POEM can be used to train the encoder to extract the day/night-invariant features by utilizing the day/night-classifying features.

\section{Acknowledgements}
This work was supported by the Institute of Information \& communications Technology Planning \& Evaluation (IITP) grant funded by the Korea government (MSIT) (No. 2020-0-01336, Artificial Intelligence Graduate School Program (UNIST)), (No. 2021-0-02201, Federated Learning for Privacy Preserving Video Caching Networks), the National Research Foundation of Korea (NRF) grant funded by the Korea government (MSIT) (No. 2021R1C1C1012797), and the Settlement Research Fund (1.200101.01) of UNIST (Ulsan National Institute of Science \& Technology).

\vspace{.2em}

\bibliography{egbib}

\onecolumn
\newpage
\appendix
\section{Supplementary Material for POEM: Polarization of Embeddings for Domain-Invariant Representations}

In this supplementary material, we provide the details of the experiment setting and other empirical results, such as source domain accuracies and the learning trend of the introduced loss terms.

\section{Details of Experiment Setting}

\subsection{Experimental Environment}
All experiments are conducted by utilizing NVIDIA Quadro RTX 8000,  400GB RAM, and Xeon(R) Gold 5218R CPU @ 2.10GHz with ubuntu 20.04 with python 3.8.12, PyTorch 1.7.1, Torchvision 0.8.2, and CUDA 11.0. Our source code is partially based on the codes of DomainBed \cite{Domainbed}, SWAD \cite{SWAD} and MIRO \cite{MIRO}.

\begin{table}[H]
	\centering
	\begin{tabular}{c|c|ccccc}
		\toprule
		\textbf{Method} & \textbf{Hyperparameter} & PACS & VLCS & OfficeHome & TerraIncognita & DomainNet \\
		\midrule
		POEM & Learning rate & 5e-5 & 1e-5 & 1e-5 & 3e-5 & 3e-5\\
		 & Dropout ratio & 0.1 & 0.1 & 0.5 & 0.5 & 0.5\\		
		 & Weight decay & 1e-6 & 1e-4 & 1e-4 & 1e-4 & 1e-6\\
		\midrule
		SWAD$^\dagger$ + POEM & Learning rate & 5e-5 & 5e-5 & 3e-5 & 5e-5 & 3e-5\\
		 & Dropout ratio & 0.1 & 0.0 & 0.5 & 0.0 & 0.5\\		
		& Weight decay & 1e-6 & 0 & 1e-4 & 0 & 1e-6\\
		& $N_s$ & 3 & 3 & 3 & 3 & 3\\
		& $N_e$ & 6 & 6 & 6 & 6 & 6\\
		& $r$ & 1.3 & 1.2 & 1.3 & 1.3 & 1.3\\		
		\midrule
		MIRO$^\dagger$ + POEM & Learning rate & 3e-5 & 1e-5 & 1e-5 & 3e-5 & 3e-5\\
		& Dropout ratio & 0.1 & 0.1 & 0.5 & 0 & 0.1\\		
		& Weight decay & 1e-4 & 1e-4 & 1e-4 & 1e-4 & 1e-6\\
		& $\lambda$ for category embedding & 0.01 & 0.01 & 0.1 & 0.1 & 0.1\\
		& $\lambda$ for domain embedding & 0.01 & 0 & 0.01 & 0 & 0\\	
		\midrule
		MIRO + SWAD$^\dagger$ + POEM & Learning rate & 3e-5 & 1e-5 & 1e-5 & 3e-5 & 3e-5\\
		& Dropout ratio & 0.1 & 0.1 & 0.5 & 0 & 0.1\\		
		& Weight decay & 1e-4 & 1e-4 & 1e-4 & 1e-4 & 1e-6\\
		& $N_s$ & 3 & 3 & 3 & 3 & 3\\
		& $N_e$ & 6 & 6 & 6 & 6 & 6\\
		& $r$ & 1.3 & 1.2 & 1.3 & 1.3 & 1.3\\
		& $\lambda$ for category embedding & 0.01 & 0.01 & 0.1 & 0.1 & 0.1\\
		& $\lambda$ for domain embedding & 0.01 & 0 & 0.01 & 0 & 0\\	
		\bottomrule
	\end{tabular}
        \footnotesize{\\$^\dagger$ indicates our implementation}\\
        \caption{Hyperparameter setting for each algorithm and dataset}
	\label{tab:HP}
\end{table}	

\subsection{Details of Hyperparameter Settings}
For all benchmarks, we search hyperparameters of the learning rate, the dropout ratio, and the weight decay for both domain-classifying and category-classifying embeddings. They are grid searched in [1e-5, 3e-5, 5e-5], [0.0, 0.1, 0.5], [1e-4, 1e-6], respectively. 
Also, when combined with SWAD, the patient parameter $N_s$, the overfitting patient parameter $N_e$, and the tolerance rate $r$ are set to be 3, 6, and 1.3, respectively. In the case of VLCS, the tolerance rate $r$ is particularly set to be 1.2, following the original setting of SWAD. 
When combining MIRO, the regularization coefficient $\lambda$ is set to be 0.1 for the PACS and VLCS experiments and 0.01 for other cases by following the original setting of MIRO. 
The regularization coefficient for domain-classifying embedding in conjunction with MIRO has been grid searched in [0, 0.1, 0.01] for each dataset. 
When $\lambda$ is zero, it indicates that domain-classifying embedding is set to be ERM. Table \ref{tab:HP} shows the chosen hyperparameter setting for all benchmarks and methods.

\section{Details of Experimental Results}
\subsection{Target Domain Accuracies}
The performance on the target domains provided in the main paper shows the averaged accuracies across target domains for each benchmark over three trials. 
The target accuracies on all test domains are shown in the tables from Table \ref{tab:PACS_detail} to Table \ref{tab:DN_detail}.

\begin{table}[H]
	\centering
	\begin{tabular}{ccccc|c}
		\toprule
		\textbf{Method} & Art painting & Cartoon & Photo & Sketch & Average\\
		\midrule\
		POEM & 85.34 $\pm$ 2.28 & 82.16 $\pm$ 0.77 & 97.01 $\pm$ 0.16 & 82.41 $\pm$ 1.14 & 86.73 \\
		SWAD$^\dagger$ + POEM & 90.12 $\pm$ 0.51 & 83.62 $\pm$ 0.33 & 97.78 $\pm$ 0.05 & 82.60 $\pm$ 0.81 & 88.53 \\
		MIRO$^\dagger$ + POEM& 87.29 $\pm$ 1.22 & 81.50 $\pm$ 0.74 & 97.65 $\pm$ 0.24 & 80.30 $\pm$ 0.03 & 86.68 \\
		MIRO + SWAD$^\dagger$ + POEM& 89.35 $\pm$ 0.70 & 83.02 $\pm$ 0.40 & 98.20 $\pm$ 0.07 & 83.21 $\pm$ 0.48 & 88.45 \\
		\hline
	\end{tabular}
	\footnotesize{\\$^\dagger$ indicates our implementation}\\
        \caption{Domain generalization accuracies on target domains in the PACS benchmark}
        \label{tab:PACS_detail}
\end{table}	

\begin{table}[H]
	\centering
	\begin{tabular}{ccccc|c}
		\toprule
		\textbf{Method} & Caltech101 & LabelMe & SUN09 & VOC2007 & Average\\
		\midrule
		POEM & 97.91 $\pm$ 0.26 & 66.71 $\pm$ 0.77 & 76.12 $\pm$ 0.48 & 76.23 $\pm$ 2.02 & 79.24 \\
		SWAD$^\dagger$ + POEM & 98.35 $\pm$ 0.24 & 64.22 $\pm$ 0.26 & 76.12 $\pm$ 0.48 & 79.01 $\pm$ 0.44 & 79.43 \\
		MIRO$^\dagger$ + POEM & 98.44 $\pm$ 0.33 & 66.44 $\pm$ 0.87 & 73.86 $\pm$ 0.83 & 77.55 $\pm$ 0.56 & 79.12 \\
		MIRO + SWAD$^\dagger$ + POEM & 98.91 $\pm$ 0.08 & 64.63 $\pm$ 0.17 & 75.63 $\pm$ 0.36 & 78.96 $\pm$ 0.46 & 79.53 \\
		\hline
	\end{tabular}
	\label{tab:VLCS_detail}
	\footnotesize{\\$^\dagger$ indicates our implementation}\\
        \caption{Domain generalization accuracies on target domains in the VLCS benchmark}
\end{table}	

\begin{table}[H]
	\centering
	\begin{tabular}{ccccc|c}
		\toprule
		\textbf{Method} & Art & Clipart & Product & Realworld & Average\\
		\midrule
		POEM & 64.06 $\pm$ 0.21 & 53.92 $\pm$ 0.73 & 76.21 $\pm$ 0.41 & 77.66 $\pm$ 0.40 & 67.96 \\
		SWAD$^\dagger$ + POEM & 67.08 $\pm$ 0.31 & 57.01 $\pm$ 0.36 & 78.21 $\pm$ 0.45 & 79.59 $\pm$ 0.46 & 70.47 \\
		MIRO$^\dagger$ + POEM& 69.60 $\pm$ 0.38 & 54.67 $\pm$ 0.67 & 79.39 $\pm$ 0.49 & 81.78 $\pm$ 0.25 & 71.36 \\
		MIRO + SWAD$^\dagger$ + POEM& 69.46 $\pm$ 0.23 & 55.41 $\pm$ 0.13 & 79.98 $\pm$ 0.14 & 82.03 $\pm$ 0.16 & 71.73\\
		\hline
	\end{tabular}
	\label{tab:OH_detail}
	\footnotesize{\\$^\dagger$ indicates our implementation}\\
        \caption{Domain generalization accuracies on target domains in the OfficeHome benchmark}
\end{table}

\begin{table}[H]
	\centering
	\begin{tabular}{ccccc|c}
		\toprule
		\textbf{Method} & location100 & location38 & location43 & location46 & Average\\
		\midrule
		POEM & 59.28 $\pm$ 1.14 & 38.83 $\pm$ 1.43 & 58.93 $\pm$ 0.97 & 40.86 $\pm$ 1.36 & 49.48 \\
		SWAD$^\dagger$ + POEM & 57.80 $\pm$ 0.20 & 47.23 $\pm$ 0.89 & 58.99 $\pm$ 0.33 & 41.97 $\pm$ 0.62 & 51.50 \\
		MIRO$^\dagger$ + POEM& 57.10 $\pm$ 4.22 & 44.13 $\pm$ 0.71 & 57.16 $\pm$ 1.56 & 38.68 $\pm$ 2.05 & 49.27 \\
		MIRO + SWAD$^\dagger$ + POEM& 59.89 $\pm$ 0.72 & 45.47 $\pm$ 0.22 & 60.37 $\pm$ 0.23 & 41.10 $\pm$ 0.47 & 51.71 \\
		\hline
	\end{tabular}
	\label{tab:Terra_detail}
	\footnotesize{\\$^\dagger$ indicates our implementation}\\
        \caption{Domain generalization accuracies on target domains in the TerraIncognita benchmark}
\end{table}	

\begin{table}[H]
	\small
	\centering
	\begin{tabular}{ccccccc|c}
		\toprule
		\textbf{Method} & Clipart & Infograph & Painting & Quickdraw & Real & Sketch & Average\\
		\midrule
		POEM & 64.41 $\pm$ 0.15 & 21.43 $\pm$ 0.30 & 49.92 $\pm$ 0.28 & 13.22 $\pm$ 0.22 & 62.22 $\pm$ 0.09 & 52.97 $\pm$ 0.09 & 44.03 \\
		SWAD$^\dagger$ + POEM & 66.67 $\pm$ 0.08 & 23.49 $\pm$ 0.02 & 54.26 $\pm$ 0.06 & 15.84 $\pm$ 0.13 & 65.69 $\pm$ 0.09 & 56.47 $\pm$ 0.08 & 47.07 \\
		MIRO$^\dagger$ + POEM& 66.73 $\pm$ 0.20 & 23.44 $\pm$ 0.06 & 53.96 $\pm$ 0.07 & 15.29 $\pm$ 0.06 & 67.19 $\pm$ 0.72 & 55.44 $\pm$ 0.20 & 47.01 \\
		MIRO + SWAD$^\dagger$ + POEM& 66.72 $\pm$ 0.19 & 23.46 $\pm$ 0.05 & 54.01 $\pm$ 0.12 & 15.27 $\pm$ 0.08 & 67.88 $\pm$ 0.04 & 55.10 $\pm$ 0.14 & 47.07 \\
		\hline
	\end{tabular}
	\footnotesize{\\$^\dagger$ indicates our implementation}\\
        \caption{Domain generalization accuracies on target domains in the DomainNet benchmark}
        \label{tab:DN_detail}
\end{table}

\subsection{Source Domain Accuracies}
We also measure the performance of POEM on the source domains. For each source domain, we split the dataset into 80\% of the samples for the training set and 20\% for the validation set. 
We evaluate two performance metrics, i.e., image-category classification accuracy and image-domain classification accuracy.
When evaluating the category classification task, the elementary embedding to classify image categories is used. 
Otherwise, when evaluating the domain classification task, the secondary embedding to classify image domains is used.
The averaged accuracies of POEM for classifying image categories and domains over three trials are shown in Table \ref{tab:source_performance}. We confirm the slight performance gain for all experiment cases. It implies that the joint training of the multiple elementary embeddings via POEM does not harm the training of the individual embedding.
To provide the details of the results, the validation accuracies for category-classification and domain-classification for each source domain are shown in the tables from Table \ref{tab:category_pacs} to \ref{tab:category_dn} and Tables \ref{tab:domain_pacs} to \ref{tab:domain_dn}, respectively.

\begin{table}[H]
	\centering
	\begin{tabular}{c}
		Accuracies for classifying image categories\\
		\begin{tabular}{cccccc|c}
			\toprule
			\textbf{Method} & PACS & VLCS & OfficeHome & TerraInc & DomainNet & Average\\
			\midrule
			ERM$^\dagger$ & 96.99 $\pm$ 0.1 & 86.21 $\pm$ 0.1 & 80.38 $\pm$ 0.1 & 91.63 $\pm$ 0.1 & 60.00 $\pm$ 0.1 & 83.31\\
			\midrule
			POEM& 97.14 $\pm$ 0.1 & 86.91 $\pm$ 0.1 & 80.85 $\pm$ 0.04 & 92.16 $\pm$ 0.1 & 60.82 $\pm$ 0.1 & 83.58\\
			\midrule
		\end{tabular}\\
		Accuracies for classifying image domains\\
		\begin{tabular}{cccccc|c}
			\toprule
			\textbf{Method} & PACS & VLCS & OfficeHome & TerraInc & DomainNet & Average\\
			\midrule
			ERM$^\dagger$ & 99.02 $\pm$ 0.1 & 94.10 $\pm$ 0.1 & 85.13 $\pm$ 0.1 & 99.95 $\pm$ 0.02& 89.26 $\pm$ 0.1 & 93.49\\
			\midrule
			POEM & 98.98 $\pm$ 0.1 & 93.99 $\pm$ 0.1 & 85.53 $\pm$ 0.1 & 99.96 $\pm$ 0.02& 89.32 $\pm$ 0.1 & 93.56\\
			\hline
		\end{tabular}
	\end{tabular}
	\footnotesize{\\$^\dagger$ indicates our implementation}\\
        \caption{Averaged validation accuracies on source domains for each benchmark}
        \label{tab:source_performance}
\end{table}

		\begin{table}[H]
	\centering
	\begin{tabular}{ccccc|c}
		\toprule
		\textbf{Method} & Art painting & Cartoon & Photo & Sketch & Average\\
		\midrule
		ERM$^\dagger$ & 97.61 $\pm$ 0.1 & 96.82 $\pm$ 0.1 & 96.29 $\pm$ 0.1 & 97.24 $\pm$ 0.2 & 96.99 $\pm$ 0.1\\
		POEM & 97.67 $\pm$ 0.1 & 96.73 $\pm$ 0.1 & 96.29 $\pm$ 0.1 & 97.86 $\pm$ 0.2 & 97.14 $\pm$ 0.1\\
		\hline
	\end{tabular}
	\footnotesize{\\$^\dagger$ indicates our implementation}\\
        \caption{Validation accuracies for image-category classification on source domains of PACS benchmark}
        \label{tab:category_pacs}
\end{table}	

\begin{table}[H]
	\centering
	\begin{tabular}{ccccc|c}
		\toprule
		\textbf{Method} & Caltech101 & LabelMe & SUN09 & VOC2007 & Average\\
		\midrule
		ERM$^\dagger$ & 81.31 $\pm$ 0.1 & 90.79 $\pm$ 0.1 & 87.33 $\pm$ 0.1 & 85.43 $\pm$ 0.1 & 86.21 $\pm$ 0.1\\
		POEM & 81.66 $\pm$ 0.2 & 91.30 $\pm$ 0.2 & 88.03 $\pm$ 0.1 & 86.69 $\pm$ 0.1 & 86.92 $\pm$ 0.1\\
		\hline
	\end{tabular}
	\label{tab:category_vlcs}
	\footnotesize{\\$^\dagger$ indicates our implementation}\\
        \caption{Validation accuracies for image-category classification on source domains of VLCS benchmark}
\end{table}	

\begin{table}[!htpb]
	\centering
	\begin{tabular}{ccccc|c}
		\toprule
		\textbf{Method} & Art & Clipart & Product & Realworld & Average\\
		\midrule
		ERM$^\dagger$ & 83.68 $\pm$ 0.2 & 81.18 $\pm$ 0.3 & 77.48 $\pm$ 0.1 & 79.17 $\pm$ 0.2 & 80.38 $\pm$ 0.1\\
		POEM & 84.23 $\pm$ 0.1 & 80.83 $\pm$ 0.2 & 77.56 $\pm$ 0.1 & 80.77 $\pm$ 0.3 & 80.85 $\pm$ 0.1\\
		\hline
	\end{tabular}
	\label{tab:category_oh}
	\footnotesize{\\$^\dagger$ indicates our implementation}\\
        \caption{Validation accuracies for image-category classification on source domains of OfficeHome benchmark}
\end{table}

\begin{table}[H]
	\centering
	\begin{tabular}{ccccc|c}
		\toprule
		\textbf{Method} & location100 & location38 & location43 & location46 & Average\\
		\midrule
		ERM$^\dagger$ & 90.40 $\pm$ 0.1 & 91.98 $\pm$ 0.1 & 91.06 $\pm$ 0.1 & 93.08 $\pm$ 0.1 & 91.63 $\pm$ 0.1\\
		POEM & 90.79 $\pm$ 0.1 & 92.20 $\pm$ 0.1 & 92.00 $\pm$ 0.1 & 93.66 $\pm$ 0.1 & 92.16 $\pm$ 0.1\\
		\hline
	\end{tabular}
	\label{tab:category_terra}
	\footnotesize{\\$^\dagger$ indicates our implementation}\\
        \caption{Validation accuracies for image-category classification on source domains of TerraIncognita benchmark}
\end{table}

\begin{table}[H]
	\centering
	\begin{tabular}{ccccccc|c}
		\toprule
		\textbf{Method} & Clipart & Infograph & Painting & Quickdraw & Real & Sketch & Average\\
		\midrule
		ERM$^\dagger$ & 57.89 $\pm$ 0.1 & 65.35 $\pm$ 0.1 & 59.90 $\pm$ 0.1 & 61.27 $\pm$ 0.1 & 56.84 $\pm$ 0.1 & 58.77 $\pm$ 0.1 & 60.00 $\pm$ 0.1 \\
		POEM & 58.04 $\pm$ 0.1 & 66.22 $\pm$ 0.1 & 60.79 $\pm$ 0.1 & 61.40 $\pm$ 0.1 & 58.52 $\pm$ 0.1 & 59.94 $\pm$ 0.1 & 60.82 $\pm$ 0.1 \\
		\hline
	\end{tabular}
	\footnotesize{\\$^\dagger$ indicates our implementation}\\
        \caption{Validation accuracies for image-category classification on source domains of DomainNet benchmark}
        \label{tab:category_dn}
\end{table}	

\begin{table}[H]
	\centering
	\begin{tabular}{ccccc|c}
		\toprule
		\textbf{Method} & Art painting & Cartoon & Photo & Sketch & Average\\
		\midrule
		ERM$^\dagger$ & 99.58 $\pm$ 0.1 & 99.07 $\pm$ 0.1 & 99.63 $\pm$ 0.2 & 97.81 $\pm$ 0.1 & 99.02 $\pm$ 0.1\\
		POEM & 99.52 $\pm$ 0.1 & 98.91 $\pm$ 0.1 & 99.27 $\pm$ 0.1 & 98.21 $\pm$ 0.1 & 98.98 $\pm$ 0.1\\
		\hline
	\end{tabular}
	\footnotesize{\\$^\dagger$ indicates our implementation}\\
        \caption{Validation accuracies for image-domain classification on source domains of PACS benchmark}
        \label{tab:domain_pacs}
\end{table}	

\begin{table}[H]
	\centering
	\begin{tabular}{ccccc|c}
		\toprule
		\textbf{Method} & Caltech101 & LabelMe & SUN09 & VOC2007 & Average\\
		\midrule
		ERM$^\dagger$ & 89.56 $\pm$ 0.2 & 93.69 $\pm$ 0.1 & 97.16 $\pm$ 0.1 & 96.00 $\pm$ 0.1 & 94.10 $\pm$ 0.1\\
		POEM & 89.19 $\pm$ 0.1 & 93.52 $\pm$ 0.1 & 97.04 $\pm$ 0.1 & 96.19 $\pm$ 0.1 & 93.99 $\pm$ 0.1\\
		\hline
	\end{tabular}
	\label{tab:domain_vlcs}
	\footnotesize{\\$^\dagger$ indicates our implementation}\\
        \caption{Validation accuracies for image-domain classification on source domains of VLCS benchmark}
\end{table}	

\begin{table}[H]
	\centering
	\begin{tabular}{ccccc|c}
		\toprule
		\textbf{Method} & Art & Clipart & Product & Realworld & Average\\
		\midrule
		ERM$^\dagger$ & 87.80 $\pm$ 0.2 & 76.49 $\pm$ 0.1 & 82.27 $\pm$ 0.1 & 93.97 $\pm$ 0.1 & 85.13 $\pm$ 0.1\\
		POEM & 88.34 $\pm$ 0.1 & 76.29 $\pm$ 0.1 & 83.17 $\pm$ 0.1 & 94.31 $\pm$ 0.1 & 85.53 $\pm$ 0.1\\
		\hline
	\end{tabular}
	\label{tab:domain_oh}
	\footnotesize{\\$^\dagger$ indicates our implementation}\\
        \caption{Validation accuracies for image-domain classification on source domains of OfficeHome benchmark}
\end{table}

\begin{table}[H]
	\centering
	\begin{tabular}{ccccc|c}
		\toprule
		\textbf{Method} & location100 & location38 & location43 & location46 & Average\\
		\midrule
		ERM$^\dagger$ & 99.93 $\pm$ 0.0 & 100.00 $\pm$ 0.0 & 99.95 $\pm$ 0.0 & 99.93 $\pm$ 0.0 & 99.95 $\pm$ 0.0\\
		POEM & 100.00 $\pm$ 0.0 & 99.93 $\pm$ 0.0 & 99.93 $\pm$ 0.0 & 99.98 $\pm$ 0.0 & 99.96 $\pm$ 0.0\\
		\hline
	\end{tabular}
	\label{tab:domain_terra}
	\footnotesize{\\$^\dagger$ indicates our implementation}\\
        \caption{Validation accuracies for image-domain classification on source domains of TerraIncognita benchmark}
\end{table}	

\begin{table}[H]
	\centering
	\begin{tabular}{ccccccc|c}
		\toprule
		\textbf{Method} & Clipart & Infograph & Painting & Quickdraw & Real & Sketch & Average\\
		\midrule
		ERM$^\dagger$ & 90.85 $\pm$ 0.1 & 87.95 $\pm$ 0.1 & 90.79 $\pm$ 0.1 & 84.21 $\pm$ 0.3 & 91.83 $\pm$ 0.2 & 89.93 $\pm$ 0.3 & 89.26 $\pm$ 0.1\\
		POEM & 90.77 $\pm$ 0.3  & 87.92 $\pm$ 0.1  & 90.76 $\pm$ 0.2  & 84.53 $\pm$ 0.2 & 91.51 $\pm$ 0.3 & 90.43 $\pm$ 0.3 & 89.32 $\pm$ 0.1\\
		\hline
	\end{tabular}
	\footnotesize{\\$^\dagger$ indicates our implementation}\\
        \caption{Validation accuracies for image-domain classification on source domains of DomainNet benchmark}
        \label{tab:domain_dn}
\end{table}	

\subsection{Learning Trend of Disentangling and Discrimination Loss Terms}
The learning trend of the key loss terms, which are the disentangling loss (or similarity loss) $\mathcal{L}_{s}$ and the discrimination loss $\mathcal{L}_{d}$ are illustrated in Fig. \ref{fig:Training_Loss_of_POEM}. As presented in Fig. \ref{fig:loss_s}, the averaged cosine similarity between the category embedding and the domain embedding over VLCS domains decreases rapidly (See the solid line colored by red). 
It means that the feature vectors from the two elementary embeddings become orthogonal. In contrast, when we drop the similarity loss term in the training of POEM, the averaged cosine similarity between elementary embeddings over VLCS domains is not zero-forced (See the dotted line colored by blue). 
It shows that the orthogonality between the features from image- and domain-classifying elementary embeddings are not trivially achieved without POEM's similarity loss. 
Fig. \ref{fig:loss_d} shows the learning trend of the discrimination loss for the five benchmarks. 
The suppressed loss values mean that the features from different embeddings become distinctive so that POEM can discriminate the features from different embeddings. 

\begin{figure*}
	\captionsetup{justification=centering}
	\centering
	\begin{subfigure}[hbt]{0.48\textwidth}
		\centering
		\includegraphics[width=\textwidth]{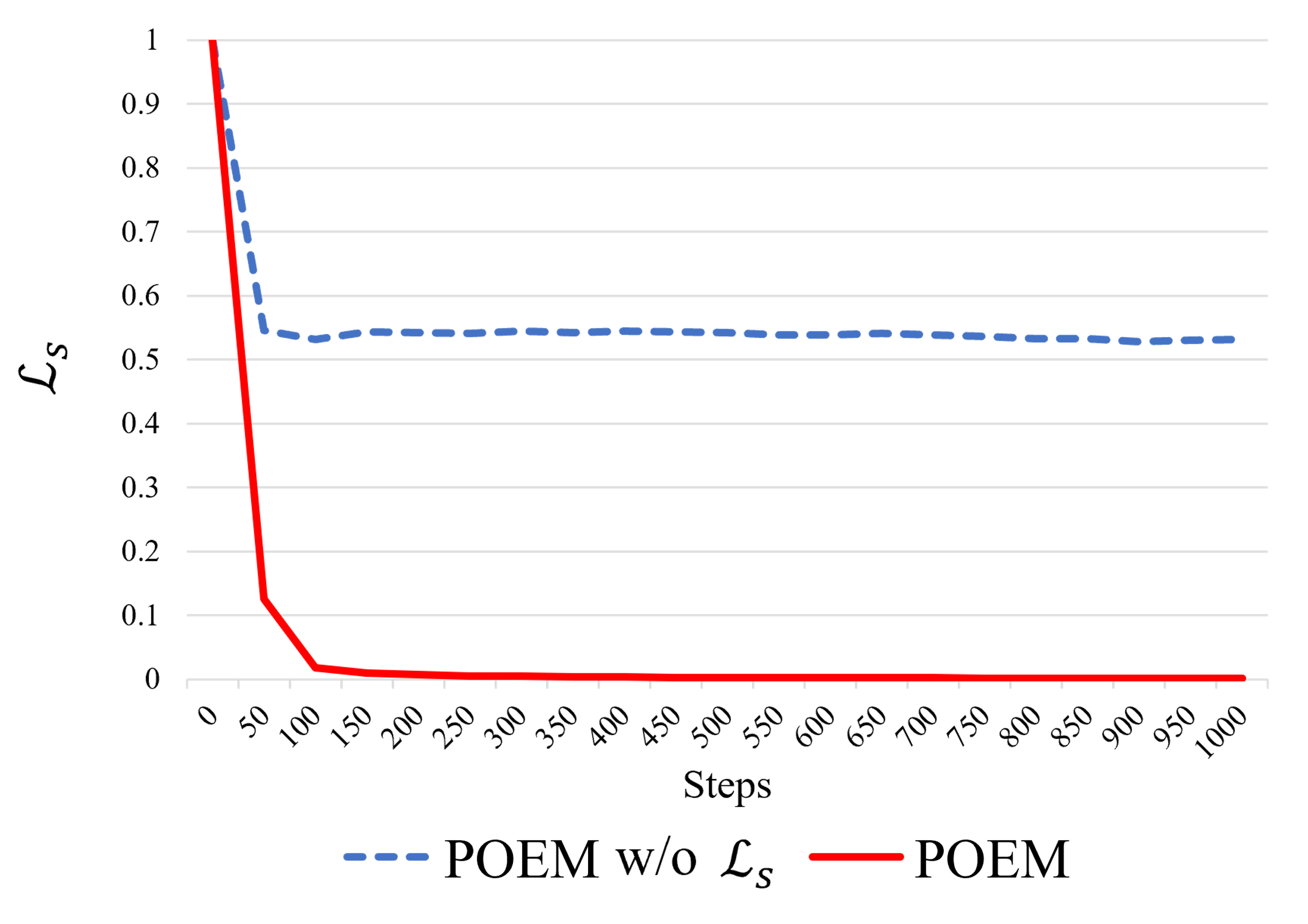}
		\caption{Trainig similarity loss $\mathcal{L}_s$}
		\label{fig:loss_s}
	\end{subfigure}
	\begin{subfigure}[hbt]{0.48\textwidth}
		\centering
		\includegraphics[width=\textwidth]{resources/task_loss/task_loss.png}
		\caption{Training discriminator loss $L_d$}
		\label{fig:loss_d}
	\end{subfigure}
	\caption{Learning trend of the component losses of POEM}
	\label{fig:Training_Loss_of_POEM}
\end{figure*}

	\subsection{Cross-Entropy between Category Features and Domain Labels}
In the main paper, we show the cross-entropy values when the category-classifying features are used to estimate the source domain.
The result emphasizes the POEM's category-classification feature is not effective in classifying domains, i.e., it implies the domain-invariance of features.
Here, we attach all calculated cross-entropy values. Every experiment with ERM and POEM uses the same random seed so that the validation set is equal. From Tables \ref{tab:CE_PACS} to \ref{tab:CE_DN}, we show the cross-entropy values. In most cases, it appears that POEM shows larger cross-entropy values than ERM which indicates that POEM learns a more domain-invariant category-classifying elementary embedding than ERM.

\begin{table}[H]
	\centering
	\begin{tabular}{ccccc|c}
		\toprule
		\textbf{Method} & Art painting & Cartoon & Photo & Sketch & Average\\
		\midrule
		ERM$^\dagger$ & 2.58 & 2.31 & 1.77 & 3.19 & 2.65\\
		POEM & 3.26 & 2.88 & 1.81 & 3.98 & 2.98 \\
		\hline
	\end{tabular}
	\footnotesize{\\$^\dagger$ indicates our implementation}\\
        \caption{Cross-Entropy for source domain data samples on PACS benchmark}
        \label{tab:CE_PACS}
\end{table}	

\begin{table}[H]
	\centering
	\begin{tabular}{ccccc|c}
		\toprule
		\textbf{Method} & Caltech101 & LabelMe & SUN09 & VOC2007 & Average\\
		\midrule
		ERM$^\dagger$ & 2.53 & 1.77 & 2.94 & 3.97 & 2.80\\
		POEM & 4.46 & 2.51 & 4.06 & 5.01 & 4.01\\
		\hline
	\end{tabular}
	\label{tab:CE_VLCS}
	\footnotesize{\\$^\dagger$ indicates our implementation}\\
        \caption{Cross-Entropy for source domain data samples on VLCS benchmark}
\end{table}	

\begin{table}[H]
	\centering
	\begin{tabular}{ccccc|c}
		\toprule
		\textbf{Method} & Art & Clipart & Product & Realworld & Average\\
		\midrule
		ERM$^\dagger$ & 1.02 & 1.21 & 1.15 & 1.13 & 1.13\\
		POEM & 1.18 & 1.36 & 1.35 & 1.76 & 1.41\\
		\hline
	\end{tabular}
	\label{tab:CE_OH}
	\footnotesize{\\$^\dagger$ indicates our implementation}\\
        \caption{Cross-Entropy for source domain data samples on OfficeHome benchmark}
\end{table}

\begin{table}[H]
	\centering
	\begin{tabular}{ccccc|c}
		\toprule
		\textbf{Method} & location100 & location38 & location43 & location46 & Average\\
		\midrule
		ERM$^\dagger$ & 1.27 & 1.98 & 2.29 & 1.86 & 1.85\\
		POEM & 1.42 & 3.31 & 2.32 & 1.44 & 2.12\\
		\hline
	\end{tabular}
	\label{tab:CE_Terra}
	\footnotesize{\\$^\dagger$ indicates our implementation}\\
        \caption{Cross-Entropy for source domain data samples on TerraIncognita benchmark}
\end{table}	

\begin{table}[H]
	\centering
	\begin{tabular}{ccccccc|c}
		\toprule
		\textbf{Method} & Clipart & Infograph & Painting & Quickdraw & Real & Sketch & Average\\
		\midrule
		ERM$^\dagger$ & 0.71 & 0.91 & 0.72 & 1.16 & 0.59 & 0.78 & 0.81 \\
		POEM & 0.59 & 0.71 & 0.58 & 1.08 & 0.48 & 0.63 & 0.68 \\
		\hline
	\end{tabular}
	\footnotesize{\\$^\dagger$ indicates our implementation}\\
        \caption{Cross-Entropy for source domain data samples on DomainNet benchmark}
        \label{tab:CE_DN}
\end{table}

\end{document}